\documentclass[10pt,twocolumn,letterpaper]{article}

\usepackage{cvpr}              % To produce the CAMERA-READY version
\usepackage{multirow,booktabs,tabularx,colortbl,xcolor}
\usepackage[hidelinks]{hyperref}
\definecolor{headergray}{RGB}{240,240,240}
\definecolor{rowlight}{RGB}{250,250,250}
\definecolor{rowdark}{RGB}{245,245,245}

\usepackage{multirow,booktabs,tabularx,colortbl,xcolor}
 \definecolor{lightgreen}{rgb}{0,0.9,0}
\definecolor{lightmintbg}{rgb}{.69,.86,.99}
\definecolor{lightmintgreen}{rgb}{0.8,0.98,0.81}
\definecolor{lightyellow}{rgb}{.99,.99,.85}
\definecolor{Grey}{gray}{0.5}

\definecolor{cvprblue}{rgb}{0.21,0.49,0.74}
\def\paperID{13209} % *** Enter the Paper ID here
\def\confName{CVPR}
\def\confYear{2026}
\usepackage{multirow,booktabs,tabularx,colortbl,xcolor}
\usepackage[hidelinks]{hyperref}
\usepackage{multirow,booktabs,tabularx,colortbl,xcolor}
\usepackage{colortbl}
\usepackage{xcolor}
\definecolor{darkgreen}{RGB}{0,120,0}
\definecolor{headergray}{RGB}{240,240,240}
\definecolor{rowlight}{RGB}{250,250,250}
\definecolor{rowdark}{RGB}{245,245,245}
\definecolor{lightgreen}{rgb}{0,0.9,0}
\definecolor{lightmintbg}{rgb}{.69,.86,.99}
\definecolor{lightmintgreen}{rgb}{0.8,0.98,0.81}
\definecolor{lightyellow}{rgb}{.99,.99,.85}
\definecolor{Grey}{gray}{0.5}
\usepackage{pifont}
\newcommand{\cmark}{\ding{51}} % check mark
\newcommand{\xmark}{\ding{55}} % cross mark
\usepackage{algorithm}
 \usepackage[numbers]{natbib}
\usepackage{amsmath,amssymb}
\usepackage{algpseudocode}
\usepackage{placeins}
\title{Annotation-Free Class-Incremental Learning}

\author{Hari Chandana Kuchibhotla$^{\ddag}$\thanks{equal contribution} , K S Ananth$^{\ddag}$\footnotemark[1] , Vineeth N Balasubramanian$^{\ddag \diamond}$\\
\normalsize $^{\ddag}$ Indian Institute of Technology Hyderabad, India $^{\diamond}$ Microsoft Research India  \\
{\tt\small \{ai20resch11006, cs22btech11029, vineethnb\}@iith.ac.in, vineeth.nb@microsoft.com}
}

\begin{document}
\maketitle
\begin{abstract}
% Despite significant progress in continual learning—ranging from architectural innovations to sophisticated strategies for mitigating catastrophic forgetting—most existing methods rest on a strong but unrealistic assumption: the availability of labeled data throughout the learning process. In real-world scenarios, however, data often arrives sequentially and without annotations, rendering conventional approaches impractical. In this work, we revisit the fundamental assumptions of continual learning and ask: Can current systems adapt when labels are absent and tasks emerge incrementally over time? To this end, we introduce Annotation-Free Class-Incremental Learning (AF-CIL), a more realistic and challenging paradigm where unlabeled data arrives continuously, and the learner must incrementally acquire new classes without any supervision. We further propose an exemplar-free, knowledge-guided framework that leverages world knowledge to generate high-quality pseudo-labels and mitigate catastrophic forgetting in a unified manner. Extensive experiments across multiple benchmarks demonstrate the inherent difficulty of the proposed setting and highlight the strong performance and adaptability of our method, achieving substantial gains over existing baselines.

Despite significant progress in continual learning ranging from architectural novelty to clever strategies for mitigating catastrophic forgetting; most existing methods rest on a strong but unrealistic assumption: "the availability of labeled data throughout the learning process". In real-world scenarios, however, data often arrives sequentially and without annotations, rendering conventional approaches impractical. In this work, we revisit the fundamental assumptions of continual learning and ask: Can current systems adapt when labels are absent and tasks emerge incrementally over time? To this end, we introduce Annotation-Free Class-Incremental Learning (AF-CIL), a more realistic and challenging paradigm where unlabeled data arrives continuously, and the learner must incrementally acquire new classes without any supervision. To enable effective learning under AF-CIL, we propose CrossWorld-CL, a Cross Domain World Guided Continual Learning framework that incorporates external world knowledge as a stable auxiliary source. The method retrieves semantically related ImageNet classes for each downstream category, maps downstream and ImageNet features through a cross-domain alignment strategy and finally introduce a novel replay strategy. This design lets the model uncover semantic structure without annotations while keeping earlier knowledge intact. Across four datasets, CrossWorld-CL surpasses CLIP baselines and existing continual and unlabeled learning methods, underscoring the benefit of world knowledge for annotation-free continual learning.

\end{abstract}    
\section{Introduction}
\label{sec:intro}

Continual Learning~\cite{gift, rapf, moeadapter, l2p} has evolved as a powerful paradigm to enable models to learn from a stream of tasks without retraining from scratch. Over the years, research in CL has largely revolved around mitigating catastrophic forgetting, the tendency of neural networks to overwrite previously learned knowledge when exposed to new data. While architectural innovations~\cite{moeadapter}, rehearsal buffers~\cite{rapf}, and regularization strategies~\cite{gift} have led to notable progress, the majority of these methods are built upon a strong but often unrealistic assumption: \textit{that labeled data is always available for every incoming task}. In reality, continual data annotation is expensive, time-consuming, and cognitively demanding, placing an unsustainable burden on human annotators. Moreover, handling unlabeled data is inherently challenging due to distributional shifts between sequentially arriving data streams and test distributions. Maintaining accuracies across tasks with such partial information where only training images and labels are available but are not paired is a challenge by itself. For example, online retail platforms continually receive new product images from vendors, many of which arrive without labels but still need to be identified and assigned to the correct category. As this process repeats, both the volume of data and the number of categories grow over time. In addition to this, with the rapidly growing privacy concerns, exemplar-based methods violate this condition. Existing methods fail miserably at addressing these challenges. To address these issues, we introduce the much needed-\textbf{\textit{Annotation-Free Class Incremental learning}} paradigm that can (i) learn effectively from unlabeled, sequentially arriving data, (ii) maintain high inference accuracy on both old and new classes, and (iii) prevent forgetting without relying on stored exemplars. \\

Traditional continual unsupervised representation learning frameworks aim to preserve the quality of learned features as unlabeled data arrives over time. These approaches often rely on contrastive or clustering-based self supervision methods such as SimCLR~\cite{simclr}, BYOL~\cite{byol}, MoCo~\cite{moco}, and DINO~\cite{dino}, combined with strategies like momentum updates or replay buffers to reduce forgetting. However, they work entirely within visual feature spaces, usually using ResNet~\cite{resnet} backbones, and therefore struggle to adapt when the underlying semantic structure of tasks changes. Their representations remain limited because they do not incorporate language or broader world knowledge.

To address these challenge, we propose \textit{CrossWorld CL}, a Cross Domain World Guided Continual Learning framework that uses external world knowledge as an auxiliary supervision.  We rely on the broad visual corpora that can encode general semantic structure, which can guide learning when annotations are missing. A proxy equivalent of world knowledge that can be publicly accessible is the widely used ImageNet\cite{imagenet} dataset due to its rich visual variability and well formed category boundaries that is otherwise missing in the unlabeled stream. 

Our proposed approach exploits the latent semantic structure between the downstream unlabeled classes and the ImageNet classes to retrieve semantically aligned auxiliary data under Task-aware World Knowledge Distillation stage. This retrieved data acts as proxy supervision that guides the model through a cross domain loss, helping it bridge the gap between unlabeled and labeled domains while also avoiding privacy concerns. While dealing with the downstream data, motivated by prior work~\cite{nola}, we make use of GPT3~\cite{gpt3} generated text descriptions for all downstream classes. Since averaged text embeddings are known to capture semantics more effectively than individual descriptions, we employ CLIP~\cite{clip} to obtain averaged text embeddings, which serve as the initial textual features under Semantic Expansion through LLM stage. These embeddings are then trained to align with the visual features within our framework. To enhance the visual supervision, we employ DINO~\cite{dino}, which is known for its strong visual backbone, to generate pseudo labels that guide the CLIP~\cite{clip} model during training. A mapping mechanism using both downstream data and the retrieved ImageNet data is designed between the DINO and CLIP feature spaces to enable smooth knowledge transfer under Dual-Supervised Visual-Semantic Alignment stage. Once DINO~\cite{dino} is trained to predict reliable pseudo labels, we integrate external world knowledge into training a prompts infused CLIP visual model using cross-domain alignment losses which greatly aids in learning the hidden semantics across domains and improves the label predictions under Prompt-guided Cross-domain Alignment stage. Lastly we use the world knowledge as pseudo-exemplars under Replay Strategy stage. By grounding continual learning in publicly available world knowledge instead of private exemplars, our framework achieves strong generalization and mitigates catastrophic forgetting without any risk of data leakage.  

\noindent Our main contributions are as follows:
\begin{itemize}
    \item We introduce \textit{Annotation-Free Class Incremental Learning} (AF-CIL), a new paradigm for unlabeled sequential data. 
    \item We introduce CrossWorld CL, a Cross Domain World Guided Continual Learning framework, that integrates external world knowledge and introduce a privacy preserving replay strategy using semantically related samples instead of storing task-specific exemplars to address Annotation-Free CIL. 
    \item We extend the CLIP–DINO synergy with cross domain alignment losses for better pseudo labeling and semantic alignment. 
    \item We show strong performance gains across four datasets, addressing a key gap in continual learning research. 
\end{itemize}
\section{Related Works}
\label{sec:related_work}

\textbf{Zero-Shot Recognition.} Zero-shot CLIP~\cite{clip} serves as a strong starting point for recognition without labels, and when adapted continually emerges into the ContinualCLIP~\cite{continualclip} framework. However, performance drops as the number of classes increases, since fixed text prompts struggle to separate fine grained categories and class overlap grows over time. Some level of adaptation or retraining becomes necessary to maintain stability making it not suitable for our task as it does not scale well with expanding class space and fails to maintain performance without labeled supervision.
\newline
\textbf{Unlabeled image recognition.}
LaFTer~\cite{lafter} improves zero shot classification by aligning CLIP’s text embeddings with visual clusters from unlabeled data using a contrastive loss, effectively tuning the classifier without any labels. NoLA~\cite{nola} extends this idea by combining CLIP and DINO, where DINO provides pseudo labels and CLIP’s text features serve as semantic anchors, refined through a cross space alignment loss. Both methods enhance zero shot generalization on static unlabeled datasets, but they operate in a single adaptation stage and assume fixed classes. Their lack of mechanisms for continual updates, privacy preservation, and long term knowledge retention makes them unsuitable for our sequential unlabeled learning setting.
\newline
\textbf{Continual Unsupervised Representation Learning.}
Approaches such as SimCLR~\cite{simclr}, BYOL~\cite{byol}, MoCo~\cite{moco}, and DINO~\cite{dino} maintain representations over time using self supervised objectives, but they operate solely within the visual space, typically with ResNet~\cite{resnet} based backbones. Their representations lack the flexibility and semantic richness of vision language models, making them unsuitable for capturing evolving category relationships. While they stabilize visual features, they cannot achieve meaningful class level recognition or adapt to semantic drift across tasks. In contrast, our setting demands models that align visual and textual semantics to remain consistent as new classes emerge.
\newline
\textbf{Continual Learning.}
L2P~\cite{l2p}, DualPrompt~\cite{dualprompt}, CODAPrompt~\cite{codaprompt}, MoEAdapter~\cite{moeadapter}, RAPF~\cite{rapf}, and GIFT~\cite{gift} represent recent advances in continual learning that mitigate forgetting through prompt tuning, adapter modules, or routing strategies. L2P~\cite{l2p} learns task specific prompts, DualPrompt~\cite{dualprompt} separates general and task dependent prompts, and CODAPrompt~\cite{codaprompt} organizes prompts in a hierarchical manner. MoEAdapter~\cite{moeadapter} employs expert adapters for task routing, while RAPF~\cite{rapf} and GIFT~\cite{gift} use dynamic parameter allocation and synthetic feature replay to preserve knowledge. Although these methods achieve strong results in labeled settings, they are all heavily dependent on annotated data to guide prompt or adapter optimization and to define task boundaries. Without labeled supervision or stored exemplars, their mechanisms for retaining and transferring knowledge collapse, making them unsuitable for annotation-free continual learning. 
\newline
\textbf{Test-time tuning approaches.}
TPT~\cite{tpt} adapts model during inference using entropy or consistency objectives. They handle small domain shifts but lack long term memory and semantic expansion. TDA~\cite{TDA} proposes a lightweight framework to adapt vision-language models like CLIP during inference using entropy minimization and feature alignment, without full retraining. It achieves fast and efficient adaptation to distribution shifts, but it operates only at test time and lacks mechanisms for long term continual learning or handling sequential unlabeled data. While test time training methods adapt quickly to shifts, they lack long term learning ability and cannot handle sequential unlabeled tasks.

\begin{figure}[t]
    \centering
    \includegraphics[width=1.0\linewidth]{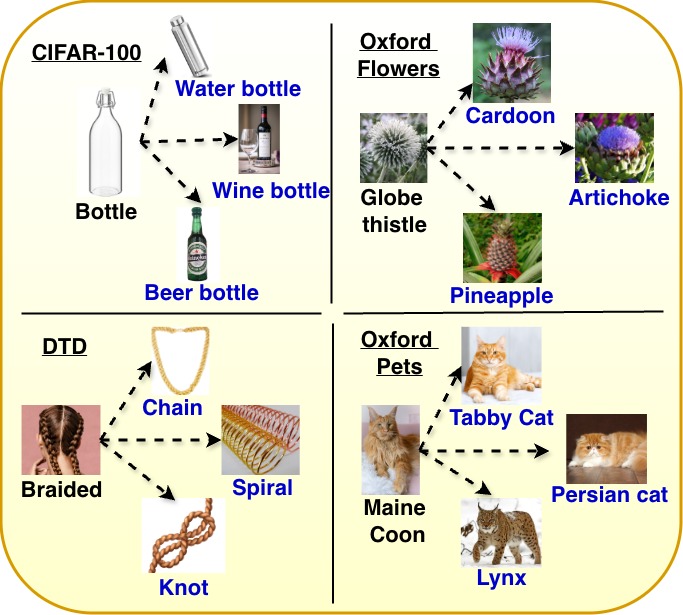}
    \caption{Examples of retrieved ImageNet classes for four downstream dataset categories; Bottle, Globe Thistle, Braided, and Maine Coon. The retrieved neighbors show clear semantic similarity to each target class, illustrating how our method gathers meaningful auxiliary supervision from ImageNet.}
    \label{fig:semantic_sim}
\end{figure}
\section{Methodology}
\label{sec:methodology}
\begin{figure*}[t]
    \centering
    \includegraphics[width=1.0\linewidth]{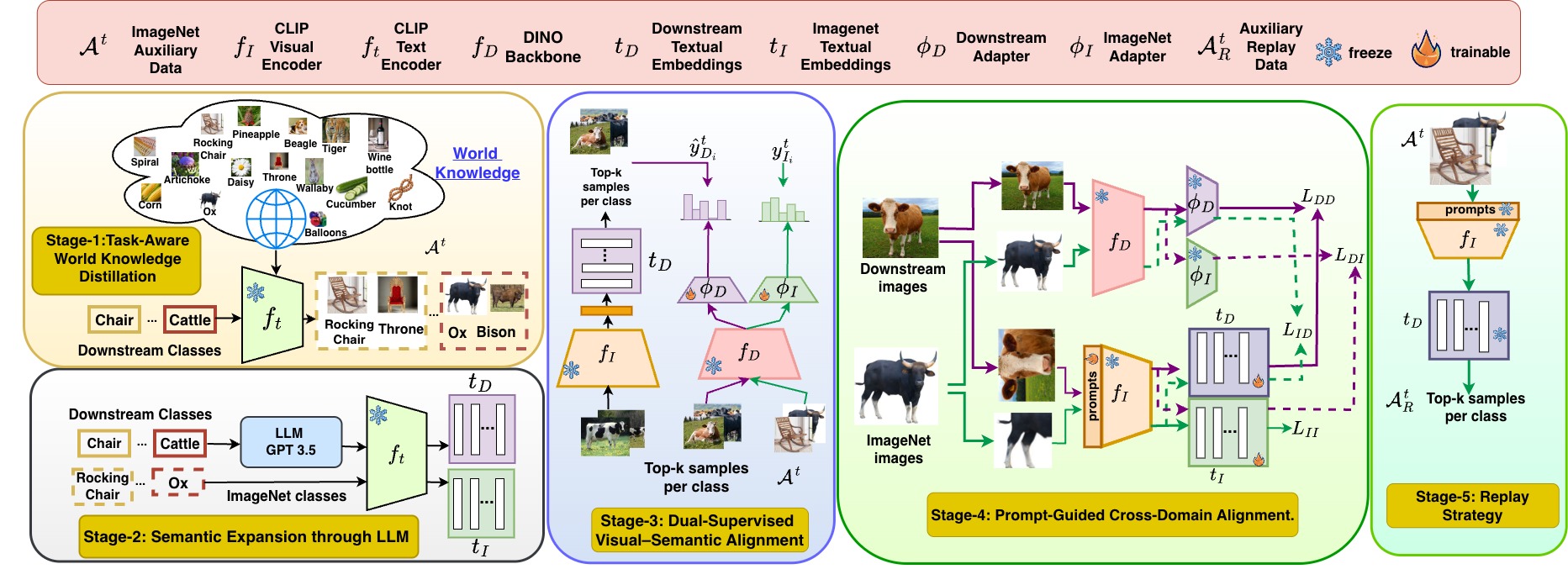}
    \caption{An overview of our annotation-free continual learning framework, which integrates world knowledge, multimodal alignment for better label prediction, and replay from public ImageNet data to learn new classes over time without the need to store downstream samples. \textcolor{violet}{\textbf{Purple}} indicates downstream data flow and \textcolor{darkgreen}{\textbf{Green}} indicates ImageNet data flow. Stage 1: Task-aware world knowledge distillation retrieves semantically related ImageNet categories for each downstream class, forming an auxiliary supervision pool $\mathcal{A}^t$. Stage 2: An LLM expands downstream labels into diverse descriptions, which are encoded with CLIP to obtain robust text prototypes for both downstream ($t_D$) and ImageNet ($t_I$) sets. Stage-3: Dual-supervised visual–semantic alignment maps DINO features into CLIP space using pseudo-labeled downstream samples and supervised ImageNet pairs. Stage-4: Prompt-guided cross-domain alignment tunes the CLIP image encoder and text prototypes using multiple alignment losses ($L_{DD}, L_{DI}, L_{ID}, L_{II}$) across downstream and ImageNet branches. Stage-5: A replay strategy constructs a lightweight exemplar set from aligned ImageNet images, enabling rehearsal without storing private downstream data $\mathcal{A}_R^t$.}
    \label{fig:main_figure}
\end{figure*}

\textbf{Preliminaries.} We begin by defining the key notations and components used in our framework, illustrated in Figure~\ref{fig:main_figure}. 
Let the unlabeled downstream data at task \( t \) be denoted as $\{ x_{D_i}^{t} \}_{i=1}^{N_D}$, representing a set of images sampled from a distribution \( \mathcal{X}_{D}^{t} \), with no labels available, where $N_D$ denotes the number of samples during task $t$. The corresponding ground truth labels, which remadin unknown during training, are denoted as $\{ y_{D_i}^{t} \}_{i=1}^{N_D}$, where $ y_{D_i}^{t} \in Y_{D}^{t}$. We further consider ImageNet dataset as our proxy for world knowledge, represented as 
$ \{ (x_{I_i}^{t}, y_{I_i}^{t}) \}_{i=1}^{N_I} $, where each $x_{I_i}^{t} \in \mathcal{X}_{I}$ is an ImageNet image and $y_{I_i}^{t} \in \mathcal{Y}_{I} $ is its corresponding semantic category label, where $N_I$ denotes the number of samples of ImageNet dataset. Subscript $D$ denotes the data is from downstream dataset and subscript $I$ denotes the data is from ImageNet dataset. At test time, the model is evaluated on the cumulative test set containing samples from all tasks seen so far, defined as $\mathcal{T}_{\text{test}}^{t} = \bigcup_{k=1}^{t} \{ (x_{\text{test}}^{k}, y_{\text{test}}^{k}) \}$ which spans the entire label space that the model has seen so far. Our framework builds upon three primary feature extractors: the CLIP text encoder $ f_{t}$, the CLIP image encoder $f_{I}$, and the DINO image encoder $f_{D}$. The objective is to leverage the semantically rich supervision from ImageNet through both its images \( x_{I_i}^{t} \) and labels \( y_{I_i}^{t} \) to generate reliable and accurate pseudo labels for the unlabeled downstream samples \( x_{D_i}^{t} \). This enables effective adaptation of the model to new classes in a continual and annotation-free manner.\\
% \vspace{3mm}
\noindent \textbf{Stage-1: Task-Aware World Knowledge Distillation.} The goal of this step is to distill task-relevant knowledge from large-scale external data such as ImageNet and use it as auxiliary supervision for the current downstream task.  
Given the set of downstream class names \( Y_{D}^{t} \) and the set of ImageNet class names $\mathcal{Y}_{I}$, we first encode all class names into the CLIP text embedding space using the text encoder $f_{t}$. We then compute the cosine similarity between each downstream class name and all ImageNet class names to identify the most semantically related ones:  
\[
\text{Sim}(c, c') = 
\frac{f_{t}(c) \cdot f_{t}(c')}{\|f_{t}(c)\|\|f_{t}(c')\|}, 
\quad c \in Y_{D}^{t}, \; c' \in \mathcal{Y}_{I}.
\]
For every downstream class $c$, we select its top-$K$ most similar ImageNet classes based on this similarity measure. The corresponding ImageNet samples and their labels  ${(x_{I_i}^{t}, y_{I_i}^{t})}$ belonging to these selected categories are then aggregated to construct an auxiliary supervised dataset:  
\[
\mathcal{A}^{t} = 
\bigcup_{c \in Y_{D}^{t}} 
\{ (x_{I_i}^{t}, y_{I_i}^{t}) \mid y_{I_i}^{t} \in \text{Top-}K(c) \}.
\]
This auxiliary dataset captures only the portion of world knowledge that is semantically relevant to the downstream task. The distilled ImageNet classes corresponding to each downstream class from four different datasets are illustrated in Figure~\ref{fig:semantic_sim}. It can be observed that the retrieved ImageNet categories are highly coherent and semantically relevant to their respective downstream classes. For instance, the downstream class "bottle" retrieves ImageNet categories such as "water bottle", "beer bottle", and "wine bottle". These examples demonstrate that the proposed semantic retrieval effectively captures both visual and conceptual similarities across domains. 
\newline

\noindent \textbf{Stage-2: Semantic Expansion through LLM.}  
In this stage, we enhance the textual representations of downstream classes by enriching them with external world knowledge. Given the label set $Y_{D}^{t}$ for task $t$, we first generate descriptive text prompts for each class using a large language model (LLM) such as GPT-3~\cite{gpt3} following ~\cite{nola}. These descriptions provide diverse semantic cues about the visual category, capturing appearance, context, and relationships with other concepts. Each textual description is then encoded using the CLIP text encoder $f_{t}$. For every class $c \in Y_{D}^{t}$, we sample $M$ textual descriptions and obtain their embeddings through $f_{t}$, followed by averaging to form a single representative text prototype:  $\mathbf{t}_{D} = \frac{1}{M} \sum_{m=1}^{M} f_{t}(p_{c}^{m})$, where $p_{c}^{m}$ denotes the $m^{\text{th}}$ textual prompt of class $c$. This averaging step yields a robust and semantically rich text feature that reduces bias from individual descriptions. For the auxiliary ImageNet dataset, we compute textual embeddings for each class label $y_{I_i}^{t} \in \mathcal{A}^{t}$ using the same CLIP text encoder $f_{t}$, and denote them as $\mathbf{t}_{I}$. The embeddings of the downstream and ImageNet classes thus lie within a shared semantic space, which facilitates effective cross-domain alignment in the later stages.
\newline
\noindent\textbf{Stage-3: Dual-Supervised Visual–Semantic Alignment.} We aim to align the DINO visual space with the CLIP semantic space by jointly utilizing the pseudo-labeled downstream data and the auxiliary ImageNet dataset. We employ DINO as the visual backbone due to its strong self-supervised learning ability and its proven effectiveness in capturing rich object-centric features without requiring labels. Unlike CLIP~\cite{clip}, which focuses on global image–text alignment, DINO~\cite{dino} learns discriminative visual representations that generalize well across unseen domains. For the downstream data samples $x_{D_i}^{t} \in X_{D}^{t}$, we first obtain pseudo labels using the CLIP image encoder $f_{I}$ and the downstream text prototypes $\mathbf{t}_{D}$. The predicted label for each sample is computed as
\[
\hat{y}_{D_i}^{t} =
\arg\max_{c \in Y_{D}^{t}}
\text{Sim}\!\left(f_{I}(x_{D_i}^{t}),\, \mathbf{t}_{D}^{c}\right).
\]
where $\text{Sim}(\cdot)$ denotes the cosine similarity between the image and text embeddings. To ensure label reliability, we select the top-$K$ samples with the highest prediction confidence for each class and keep them aside for further supervision. Next, our goal is to use both the pseudo-labeled downstream samples $\{ (x_{D_i}^{t}, \hat{y}_{D_i}^{t}) \}$ and the auxiliary supervised ImageNet data $\mathcal{A}^t$, to learn a mapping between the DINO and CLIP spaces. To achieve this, we freeze the DINO backbone $f_{D}$ and attach two lightweight adapter networks; one for the downstream data and one for the ImageNet data denoted as $\phi_{D}$ and $\phi_{I}$, respectively. Each adapter outputs logits corresponding to the number of classes in its domain, that is, $|\phi_{D}| = |Y_{D}^{t}|$ and $|\phi_{I}| = |\mathcal{Y}_{I}^t|$. The downstream adapter $\phi_{D}$ is trained using the pseudo labels $\hat{y}_{D_i}^{t}$ as supervisory signals, while the ImageNet adapter $\phi_{I}$ is trained using the available ground-truth labels $y_{I_i}^{t}$. The learning objective minimizes the cross-entropy loss across both domains:  
% $
% \mathcal{L}_{\text{map}} = 
% \mathcal{L}_{\text{CE}}(\phi_{D}(f_{d}(x_{D_i}^{t})), \hat{y}_{D_i}^{t}) + 
% \mathcal{L}_{\text{CE}}(\phi_{I}(f_{d}(x_{I_i}^{t})), y_{I_i}^{t}).
% $  
\[
\mathcal{L}_{\text{map}} =
\mathcal{L}_{\text{CE}}\!\left(\phi_{D}(f_{D}(x_{D_i}^{t})),\, \hat{y}_{D_i}^{t}\right)
+
\mathcal{L}_{\text{CE}}\!\left(\phi_{I}(f_{D}(x_{I_i}^{t})),\, y_{I_i}^{t}\right).
\]

This step allows the DINO representations to be semantically grounded in the CLIP space by leveraging both pseudo and real supervision signals. As a result, the downstream DINO features become aligned with the language-driven CLIP embeddings, enabling better generalization and improved performance in the annotation-free continual learning setting.
\begin{figure}[t]
    \centering
    \includegraphics[width=1.0\linewidth]{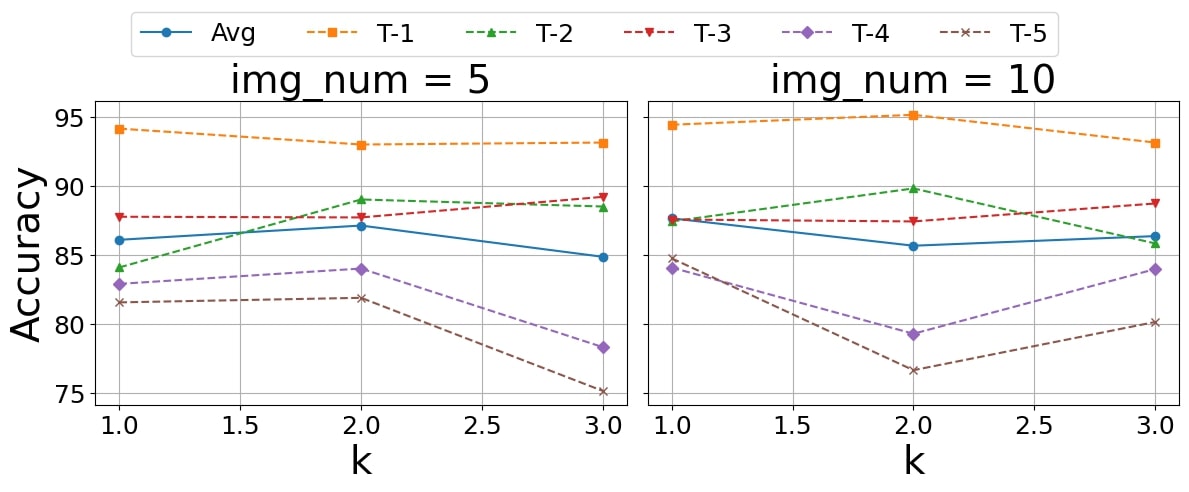}
    \caption {Effect of varying the number of retrieved ImageNet classes ($k$) and images per class (\texttt{img\_num} = 5, 10) on task-wise accuracy across all five tasks for Oxford Pets dataset. Increasing \texttt{img\_num} generally stabilizes performance, especially for later tasks Task-4 and Task-5.}
    \label{fig:acc_imgs}
\end{figure}
\newline
\noindent \textbf{Stage 4: Prompt-Guided Cross-Domain Alignment.}  
In this stage, we perform cross-domain feature alignment to unify the representations of the downstream and auxiliary ImageNet datasets within a shared semantic space. Unlike the previous stage, where DINO was aligned with CLIP through adapter-based mapping, this stage focuses on refining the CLIP space itself to better capture task-specific semantics. To achieve this, we introduce learnable visual prompts into the CLIP image encoder, resulting in a prompted encoder denoted as $f_{I_p}$. These prompts are inserted into the patch token sequence of the vision transformer and act as lightweight, learnable context tokens that allow the encoder to adapt its visual representations to the downstream task while retaining CLIP’s pretrained generalization ability. The trainable parameters in this stage are the visual prompts within $f_{I_p}$ and the textual embeddings corresponding to the downstream and ImageNet classes, namely $\mathbf{t}_{D}$ and $\mathbf{t}_{I}$. Together, these parameters enable fine-grained semantic adaptation across domains under minimal supervision. In this stage, the DINO backbone $f_{D}$ and both adapters $\phi_{D}$ and $\phi_{I}$ are frozen. They now act as reliable pseudo-label generators to guide the learning of the prompted CLIP encoder. Every downstream image $x_{D_i}^{t}$ and auxiliary ImageNet image $x_{I_i}^{t}$ undergo two augmentations; an \textit{identity transformation} and a \textit{strong augmentation}. The identity-augmented samples are passed through the DINO encoder $f_{D}$ to generate pseudo labels, while the strongly augmented samples are passed through the prompted CLIP image encoder $f_{I_p}$. The key contribution of our work lies in the proposed cross-domain alignment mechanism, where the DINO backbone features are passed through both the downstream adapter $\phi_{D}$ and the ImageNet adapter $\phi_{I}$ to generate same domain and cross-domain pseudo labels, while the prompted CLIP visual features are projected onto the corresponding text embeddings $t_D$ and $t_I$ to provide predictions. These cross-domain pseudo labels promote generalization and also serve as replay cues. The main objective to map ImageNet data into the downstream label space is for two key reasons; i) it provides an auxiliary supervisory signal that enhances semantic understanding of the downstream classes, ii) it supports privacy-preserving replay by substituting downstream exemplars with semantically aligned ImageNet samples. To achieve this, multiple alignment objectives are designed to ensure consistency across both domains (downstream and ImageNet) and modalities (image and text). The overall loss is defined as  

\[
\mathcal{L}_{\text{align}} = \mathcal{L}_{DD} +\lambda_{1} \mathcal{L}_{II} + \lambda_{2}\mathcal{L}_{ID} + \lambda_{3}\mathcal{L}_{DI},
\]
where each component contributes to a specific alignment:  
\noindent i) \textbf{Downstream-Downstream (D2D) alignment}:  
This loss uses the DINO-generated pseudo labels and predicted labels from $t_D$ to supervise the CLIP branch for downstream data:
\[
\mathcal{L}_{DD} = \mathcal{L}_{\text{CE}}(t_{D}(f_{I_p}(\textit{aug}(x_{D_i}^{t}))), \phi_D(f_D(x_{D_i}^{t}))).
\]
\noindent ii) \textbf{ImageNet-ImageNet (I2I) alignment}:
This loss uses ImageNet ground-truth labels to train the prompted CLIP features directly:
\[
\mathcal{L}_{II} = \mathcal{L}_{\text{CE}}(t_I(f_{I_p}(\textit{aug}(x_{I_i}^{t}))), y_{I_i}^{t}).
\]

\noindent iii) \textbf{ImageNet-to-Downstream (I2D) alignment}:  
This key loss term maps ImageNet features into the downstream label space through the downstream adapter:
\[
\mathcal{L}_{ID} = \mathcal{L}_{\text{CE}}(t_{D}(f_{I_p}(\textit{aug}(x_{I_i}^{t}))), \phi_{D}(f_{D}(x_{I_i}^{t}))).
\]
In this process, the DINO features of ImageNet images are passed through the downstream adapter $\phi_{D}$ to generate pseudo labels in the downstream domain, while the augmented ImageNet images are simultaneously forwarded through the prompted CLIP visual encoder $f_{I_p}$ and projected onto the downstream text embeddings $\mathbf{t}_{D}$ to predict class labels. Together, these complementary pathways establish a consistent semantic mapping from ImageNet to the downstream label space.

\noindent iv) \textbf{Downstream-to-ImageNet (D2I) alignment}: 
This complementary term encourages downstream features to align with ImageNet semantics via the ImageNet adapter:
\[
\mathcal{L}_{DI} = \mathcal{L}_{\text{CE}}( t_I(f_{I_p}(\textit{aug}(x_{D_i}^{t}))),\phi_{I}(f_{D}(x_{D_i}^{t}))).
\]
\begin{table}[t]
\centering
\renewcommand{\arraystretch}{1.2}
\setlength{\tabcolsep}{6pt}
\scalebox{0.3}{\Huge
\begin{tabular}{c|c|c|c|c|c|c|c|c|c|c|c}
\toprule
$\mathcal{L}_{DD}$ & $\mathcal{L}_{II}$ & $\mathcal{L}_{DI}$ & $\mathcal{L}_{ID}$ & $\mathcal{L}_{KL_D}$ & $\mathcal{L}_{KL_I}$ 
& Avg & T1 & T2 & T3 & T4 & T5 \\
\midrule

\cmark & \xmark & \xmark & \xmark & \cmark & \xmark
& 76.572 & 93.55 & 86.54 & 83.95 & 62.76 & 56.06 \\

\cmark & \xmark & \xmark & \xmark & \cmark & \cmark
& 78.522 & 93.55 & 86.25 & 85.63 & 67.41 & 59.77 \\

\cmark & \cmark & \xmark & \xmark & \cmark & \cmark
& 77.062 & 93.12 & 86.98 & 84.67 & 65.05 & 55.49 \\

\cmark & \xmark & \cmark & \xmark & \cmark & \cmark
& 76.772 & 93.12 & 87.34 & 83.81 & 64.70 & 54.89 \\

\cmark & \xmark & \xmark & \cmark & \cmark & \cmark
& 86.312 & 93.12 & 85.53 & 88.76 & 83.58 & 80.57 \\
\rowcolor{lightmintgreen}
\cmark & \cmark & \cmark & \cmark & \cmark & \cmark
& 86.348 & 93.12 & 85.82 & 88.71 & 83.96 & 80.13 \\

\bottomrule
\end{tabular}
}
\caption{Ablation study evaluating the contribution of each loss component in our framework for Oxford Pets dataset. The best results are obtained with all the proposed components.}
\label{tab:loss_ablation}
\end{table}
Together, these objectives ensure bidirectional supervision between the downstream and ImageNet domains, allowing the CLIP visual encoder to learn a task-aware and semantically consistent representation. 
Through prompt tuning of $f_{I_p}$ and joint refinement of $\mathbf{t}_{D}$ and $\mathbf{t}_{I}$, the model successfully bridges the DINO and CLIP spaces, achieving effective annotation-free continual learning while preserving privacy and improving cross-domain generalization.
\begin{table*}[t]
    \centering
    \scalebox{0.234}{\Huge
    \begin{tabular}{l|cccccc|cccccc|cccccc|cccccc|c}
    \toprule
    \multicolumn{1}{c|}{\textbf{Method}} & \multicolumn{6}{c|}{\textbf{DTD}} & \multicolumn{6}{c|}{\textbf{CIFAR-100}} & \multicolumn{6}{c|}{\textbf{Oxford Pets}} & \multicolumn{6}{c|}{\textbf{Oxford Flowers}} & \textbf{Overall Avg} \\
    \midrule
        & Avg & T1 & T2 & T3 & T4 & T5 
        & Avg & T1 & T2 & T3 & T4 & T5 
        & Avg & T1 & T2 & T3 & T4 & T5 
        & Avg & T1 & T2 & T3 & T4 & T5 &  \\
    \midrule

    % \multicolumn{28}{c}{\rowcolor{lightyellow}\textbf{CLIP-based}} \\
    % \midrule
    \multicolumn{1}{c|}{\rule{0pt}{25pt}\textbf{CLIP-based}\rule[-7pt]{0pt}{14pt}} & \multicolumn{6}{c|}{} & \multicolumn{6}{c|}{} & \multicolumn{6}{c|}{} & \multicolumn{6}{c|}{} & \\
    \rule{-7pt}{25pt} Continual CLIP  & 47.04 & 65.0 & 50.83 & 42.43 & 39.47 & 37.47 & 48.38 & 64.68 & 53.36 & 47.08 & 39.37 & 37.41 & 78.47 & 90.11 & 76.34 & 76.26 & 73.83 & 75.82 & 64.00 & 80.91 & 67.95 & 58.04 & 57.41 & 55.70 & 59.47 \\
    CLIP-PL  & 46.43 & 75.28 & 56.81 & 24.62 & 40.28 & 35.17 & \underline{64.46} & 93.32 & 74.92 & 56.21 & 48.48 & 49.38 & 73.94 & 94.41 & 88.86 & 82.41 & 53.35 & 50.67 & 50.00 & 88.32 & 50.50 & 42.97 & 30.94 & 37.27 & 58.21 \\
    \midrule
    % \multicolumn{28}{c}{\rowcolor{lightyellow}\textbf{{Unlabeled Image Recognition}}} \\
    % \midrule
    \multicolumn{1}{c|}{\rule{0pt}{25pt}\textbf{Unlabeled Image Recog.}\rule[-7pt]{0pt}{14pt}} & \multicolumn{6}{c|}{} & \multicolumn{6}{c|}{} & \multicolumn{6}{c|}{} & \multicolumn{6}{c|}{} & \\
    \rule{-7pt}{25pt} Seq LaFTer (CVPR’24) & 35.02 & 78.06 & 37.78 & 24.33 & 19.23 & 15.72 & 36.35 & 90.21 & 40.44 & 26.94 & 12.96 & 11.18 & 40.49 & 91.98 & 45.59 & 28.06 & 20.41 & 16.43 & 33.48 & 88.60 & 40.06 & 16.71 & 14.82 & 7.23 & 36.84 \\
    Seq NoLA (ArXiv'24) & \underline{54.54} & 86.38 & 77.08 & 50.57 & 32.89 & 25.80 & 45.12 & 93.31 & 58.56 & 26.86 & 26.57 & 20.32 & 62.80 & 94.27 & 82.52 & 53.78 & 45.81 & 37.66 & 67.12 & 94.87 & 79.68 & 58.35 & 55.68 & 47.05 & 57.90 \\
    \midrule

    % \multicolumn{27}{c}{\rowcolor{lightyellow}\textbf{Test-Time Tuning}} \\
    % \midrule
   \multicolumn{1}{c|}{\rule{0pt}{25pt}\textbf{Test-Time Tuning}\rule[-7pt]{0pt}{14pt}} & \multicolumn{6}{c|}{} & \multicolumn{6}{c|}{} & \multicolumn{6}{c|}{} & \multicolumn{6}{c|}{} & \\
    \rule{-7pt}{25pt} TPT (ICLR’23) & 25.83 & 40.28 & 23.75 & 21.07 & 21.42 & 22.64 & 58.44 & 72.21 & 66.15 & 55.94 & 50.45 & 47.49 & 69.46 & 74.64 & 66.43 & 72.66 & 65.43 & 68.17 & 30.04 & 47.86 & 28.61 & 26.42 & 25.44 & 21.88 & 45.94 \\
    \midrule

    % \multicolumn{28}{c}{\rowcolor{lightyellow}\textbf{Continual Learning}} \\
    % \midrule
    \multicolumn{1}{c|}{\rule{0pt}{25pt}\textbf{Continual Learning}\rule[-7pt]{0pt}{14pt}} & \multicolumn{6}{c|}{} & \multicolumn{6}{c|}{} & \multicolumn{6}{c|}{} & \multicolumn{6}{c|}{} & \\
    \rule{-7pt}{25pt} L2P (CVPR’22) & 23.85 & 57.50 & 31.39 & 15.71 & 9.28 & 5.38 & 23.71 & 86.95 & 10.64 & 8.32 & 8.08 & 4.58 & 31.88 & 72.78 & 38.78 & 22.68 & 15.10 & 10.06 & 23.16 & 60.40 & 29.18 & 13.75 & 5.88 & 6.37 & 25.65 \\
    DualPrompt (ECCV’22) & 31.27 & 72.78 & 35.42 & 20.02 & 15.50 & 12.65 & 23.48 & 87.68 & 14.51 & 5.74 & 5.71 & 3.79 & 36.49 & 80.80 & 40.67 & 27.10 & 19.40 & 14.50 & 31.10 & 73.50 & 37.48 & 19.89 & 14.77 & 9.87 & 30.59 \\
    CODAPrompt (CVPR’23) & 37.63 & 80.83 & 40.69 & 28.16 & 21.27 & 17.20 & 38.47 & 83.11 & 40.49 & 29.79 & 21.63 & 17.35 & 37.19 & 81.66 & 41.24 & 27.39 & 19.78 & 15.89 & 37.13 & 88.32 & 44.35 & 24.01 & 16.55 & 12.42 & 37.10 \\
    MoEAdapter (CVPR’24) & 30.61 & 88.89 & 31.81 & 16.38 & 10.31 & 5.67 & 17.52 & 38.89 & 20.90 & 11.89 & 10.89 & 5.06 & 6.49 & 14.33 & 7.16 & 4.76 & 3.47 & 2.73 & 37.99 & 89.17 & 24.46 & 40.09 & 22.53 & 13.72 & 23.15 \\
    RAPF (ECCV’24) & 53.66 & 79.17 & 57.64 & 47.32 & 43.35 & 40.84 & 58.04 & 82.89 & 63.79 & 54.06 & 46.58 & 42.92 & \underline{80.74} & 92.69 & 79.09 & 78.42 & 75.81 & 77.71 & \underline{69.03} & 91.97 & 73.68 & 61.69 & 60.32 & 57.53 & \underline{65.37} \\
    GIFT (CVPR’25) & 51.11 & 88.06 & 63.19 & 45.59 & 30.12 & 28.61 & 57.32 & 93.74 & 65.00 & 45.94 & 47.62 & 34.34 & 52.87 & 91.83 & 63.17 & 44.31 & 34.85 & 30.23 & 57.23 & 95.73 & 65.09 & 57.26 & 38.27 & 29.84 & 54.13 \\
    \midrule
    
    % \multicolumn{28}{c}{\rowcolor{lightyellow}\textbf{Ours}} \\
    % \midrule
    \multicolumn{1}{c|}{\rule{0pt}{25pt}\textbf{Ours}\rule[-7pt]{0pt}{14pt}} & \multicolumn{6}{c|}{} & \multicolumn{6}{c|}{} & \multicolumn{6}{c|}{} & \multicolumn{6}{c|}{} & \\
    \rule{-7pt}{25pt} \Huge\textbf{CrossWorld-CL}  & \cellcolor{lightmintgreen}\textbf{62.65} & 84.17 & 74.31 & 57.85 & 51.10 & 45.80 & \cellcolor{lightmintgreen}\textbf{67.44} & 93.11 & 79.85 & 66.60 & 51.45 & 46.19 & \cellcolor{lightmintgreen}\textbf{86.35} & 93.12 & 85.82 & 88.71 & 83.96 & 80.16 & \cellcolor{lightmintgreen}\textbf{71.21} & 94.02 & 80.54 & 63.95 & 60.86 & 56.68 & \cellcolor{lightmintgreen}\textbf{71.91} \\
    &\textcolor{blue}{\textbf{(+8.11)}}&&&&&&\textcolor{blue}{\textbf{(+2.98)}}&&&&&&\textcolor{blue}{\textbf{(+5.60)}}&&&&&&\textcolor{blue}{\textbf{(+2.18)}}&&&&&&\textcolor{blue}{\textbf{(+6.54)}}\\

    \bottomrule
    \end{tabular}
    }
    \caption{\textbf{Main Result:} We comparison our method with different methods namely CLIP baselines, unlabeled image recognition methods, test-time tuning methods, and continual learning approaches across four datasets: DTD, CIFAR-100, Oxford Pets, and Oxford Flowers. For each method, we report per-task accuracy (Task-1 to Task-5), dataset-level averages, and the overall average across all benchmarks. Our approach achieves the highest overall accuracy, with consistent gains on all datasets, particularly on later tasks highlighting the benefit of using auxiliary world knowledge and cross-domain alignment to improve robustness and reduce forgetting in the annotation-free continual learning setting. All results are reported in \%.}
    \label{tab:maintable}
\end{table*}

\noindent \textbf{Stage-5: Replay Strategy.} To mitigate catastrophic forgetting, after completing training at task $t$, we use the prompted CLIP image encoder $f_{I_p}$ to extract visual embeddings for the ImageNet images. Each embedding is then compared against the downstream text prototypes $\mathbf{t}_{D}$, and the top-$K$ ImageNet samples per class from $\mathcal{A}^t$ are selected based on similarity. These selected ImageNet images are associated with the corresponding downstream task labels, effectively forming a synthetic replay dataset that is publicly available, privacy-safe, and domain-independent. Because of the previously established cross-domain alignment, these samples are well-aligned semantically and can serve as reliable proxies for replay without storing any task-specific exemplars. In addition to replay, we adopt a temporal knowledge distillation (KD) strategy to stabilize the textual prototypes over time. Following each task, we store the text prototypes $\mathbf{t}_{D}^{t-1}$ and $\mathbf{t}_{I}^{t-1}$ from the previous task and regularize the current prototypes $\mathbf{t}_{D}^{t}$ and $\mathbf{t}_{I}^{t}$ to remain consistent through a Kullback–Leibler (KL) divergence loss:  
\begin{align*}
\mathcal{L}_{\text{KD}} = &
\ \text{KL}\left(
\text{Softmax}\left(\frac{\mathbf{t}_{D}^{t-1}}{\tau}\right)
\ \bigg\|\ 
\text{Softmax}\left(\frac{\mathbf{t}_{D}^{t}}{\tau}\right)
\right) \nonumber \\
& +
\text{KL}\left(
\text{Softmax}\left(\frac{\mathbf{t}_{I}^{t-1}}{\tau}\right)
\ \bigg\|\ 
\text{Softmax}\left(\frac{\mathbf{t}_{I}^{t}}{\tau}\right)
\right),
\end{align*}
\[
\mathcal{L}_{\text{total}} = \mathcal{L}_{\text{align}} + \lambda_{4}\mathcal{L}_{\text{KD}}.
\]

where $\tau$ is a temperature scaling factor that controls the smoothness of the probability distributions. This loss constrains the evolution of text prototypes across tasks, ensuring that semantic drift is minimized while still allowing adaptation to new classes.

\section{Experiments}
\label{sec:experiments}

\noindent\underline{\textbf{Baselines.}} In our experiments, we compare the proposed method against a diverse set of strong baselines spanning four major categories. \underline{CLIP-based baselines:} We compare against Continual CLIP~\cite{continualclip}, which incrementally adapts the CLIP model over tasks, and CLIP with Pseudo-Label (CLIP-PL) Training, where pseudo labels generated by CLIP are used for supervision. \underline{Unlabeled image recognition baselines:} We evaluate Seq-NoLA and Seq-LaFTer, which adapt NoLA~\cite{nola} and LaFTer~\cite{lafter} to continual learning by applying their unlabeled image tuning strategy sequentially across tasks. \underline{Test-time tuning baselines:} We evaluate on TPT ~\cite{tpt}, which adapts CLIP prompts during inference to handle distribution shifts. \underline{Continual Learning baselines:} We benchmark against several state-of-the-art prompt-based and adapter-based methods; L2P~\cite{l2p}, DualPrompt~\cite{dualprompt}, CODAPrompt~\cite{codaprompt}, MoEAdapter~\cite{moeadapter}, RAPF~\cite{rapf}, and GIFT~\cite{gift}, all of which assume labeled data but have been designed uniquely to handle catastrophic forgetting. Here, instead of labeled data, we use CLIP pseudo labels as proxy labels for the given downstream images. \underline{\textbf{Datasets:}} We conduct evaluations on four challenging datasets representing diverse visual domains which are DTD~\cite{dtd}, CIFAR-100~\cite{cifar100}, Oxford Pets~\cite{oxfordpets}, and Oxford Flowers~\cite{oxfordflowers} to assess both generalization and robustness. As there are no prior works, we evaluate on a new benchmark, where we divide the dataset into 5 tasks and distribute the total number of classes equally across 5 tasks. \underline{\textbf{Metrics:}} For evaluation, we compute classification accuracy for each task and report both the task-wise accuracies and the average accuracy across all tasks to measure overall continual performance.

\subsection{Implementation Details}
We build on CLIP ViT‑B/32 with all CLIP and DINO backbones frozen; only (i) T visual prompt tokens, (ii) CLIP LayerNorm scale/bias, and (iii) a lightweight class adapter (Linear 512$\rightarrow$C, no bias) are trainable. Class prototypes are computed by averaging L2‑normalized text embeddings from dataset‑specific prompts and “a photo of a \{\}.” used both to initialize the classifier and to retrieve up to three nearest ImageNet synsets per class. Replay stores 10 images per semantically matched synset and runs via a small ImageNet loader alongside downstream data. We train with AdamW (lr 0.004, betas 0.9/0.999, wd 0.01) and step‑decay 0.2; batch sizes are 256 (downstream), 32 (top‑k warm‑up), and 64 (replay, 4 workers). Losses are cross‑entropy in downstream and ImageNet spaces and  $\lambda_1$, $\lambda_2$, $\lambda_3$=1 and $\lambda_4$ = 30.
Trainable parameters are $768T + 512C + 39,936$, where $T$ is the number of prompt tokens (default 16), $C$ is the number of downstream classes, and $39{,}936$ is the number of LayerNorm parameters in CLIP ViT-B/32, i.e., for CIFAR‑100 $~103k$; Oxford Pets $~71k$; DTD $~76k$; Oxford Flowers $~104k$ which is hardly 0.047–0.069\% of the total trainable parameters (151M) of ViT-B/32.
Replay remains tiny: by Task 5 the union of ImageNet proxy classes is 122 (Pets), 87 (DTD), 219 (Flowers), 344 (CIFAR‑100), yielding only 870–3,440 images (Approx 0.00068–0.00269 fraction of the 1.28M‑image ImageNet train set). Experiments use PyTorch 2.4.1/CUDA 12.2 on a Tesla V100‑SXM2‑32GB. We will make our code publicly available upon acceptance.

\begin{figure*}
    \centering
    \includegraphics[width=1.0\linewidth]{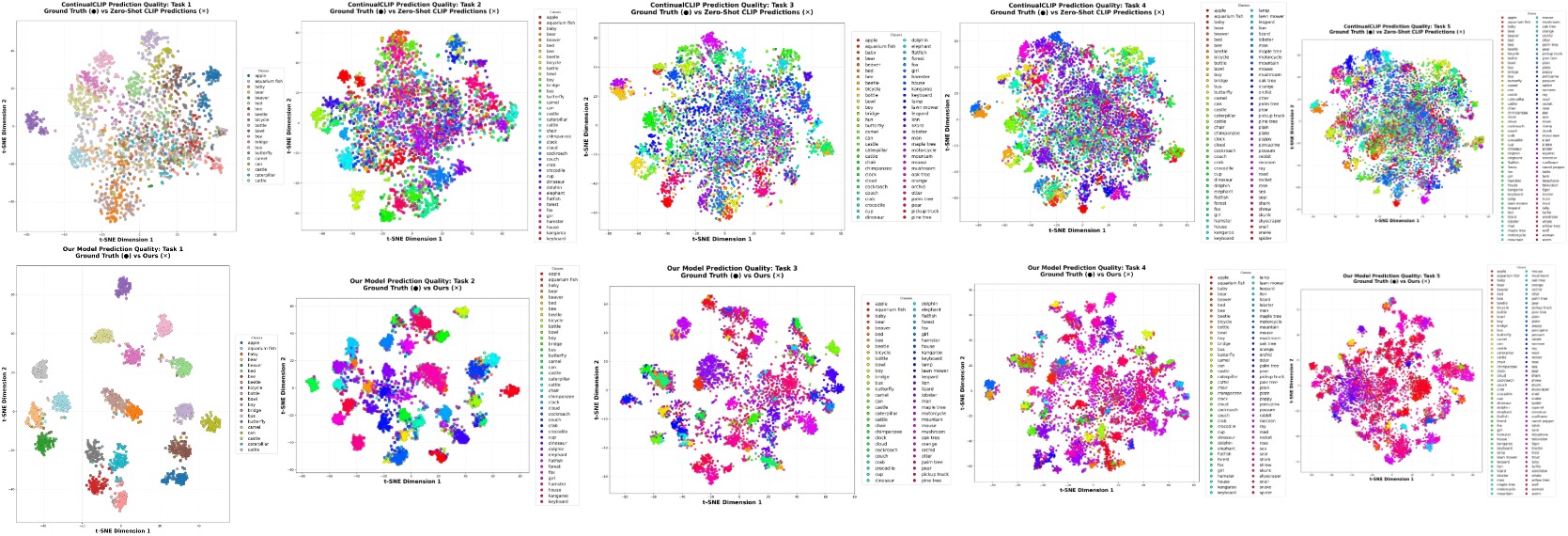}
    \caption{t\mbox{-}SNE visualizations comparing prediction quality of Continual CLIP (top row) and our method (bottom row) across all five tasks for CIFAR100 dataset. Continual CLIP produces highly entangled clusters with significant overlap between classes, indicating poor separation and increasing confusion as tasks progress. In contrast, our model yields well formed, compact, and clearly separated clusters for every task, showing that cross domain alignment and world knowledge driven supervision lead helps.}
    \label{fig:tsne}
\end{figure*}

\subsection{Main Results}
\noindent The results presented in Table ~\ref{tab:maintable}, show a consistent and significant advantage of our method across all four datasets compared to CLIP-based, Unlabeled image recognition, test-time tuning, and continual learning baselines. CLIP-only methods struggle as the number of classes grows, particularly in the later tasks (Task-3, Task-4, and Task-5), where class boundaries become finer and the visual variations between categories increase. Unlabeled image recognition baselines such as Seq NoLA and Seq LaFTer improve the early-task accuracy but degrade sharply in later tasks due to error accumulation in pseudo labeling and the lack of mechanisms to handle distribution drift. Classical continual learning approaches (L2P~\cite{l2p}, DualPrompt~\cite{dualprompt}, CODAPrompt~\cite{codaprompt}, MoEAdapter~\cite{moeadapter}, RAPF~\cite{rapf}, GIFT~\cite{gift}), although strong under supervised settings, collapse in this annotation-free scenario because they rely heavily on labeled exemplars and fail to maintain discriminative features without them. In contrast, our method achieves the highest average performance on every dataset, with especially large margins on Oxford Pets and Oxford Flowers, where fine-grained distinctions make the task more challenging. A key observation is that our gains are not limited to early tasks but become more pronounced in later tasks. This demonstrates that the auxiliary world knowledge from ImageNet stabilizes learning as new classes arrive, providing strong semantic anchors that prevent confusion among visually similar categories.

\subsection{Further Analysis}
\noindent \textbf{Ablation Study.} To better understand the contribution of each loss component, we systematically enable and disable them and study the effect on all task accuracies as shown in Table~\ref{tab:loss_ablation}. Firstly, when either of the cross-domain alignment losses ($\mathcal{L}_{DI}$ or $\mathcal{L}_{ID}$) is removed, the model loses its ability to reliably connect downstream samples with the auxiliary ImageNet supervision. This doesn't completely effect the early tasks, where the label space is still small, but it becomes much more noticeable on the later tasks starting from Task-3. The drop in accuracy in these settings suggests that the model increasingly relies on the auxiliary supervision to keep class boundaries stable as the task sequence grows. \\
\noindent Without these losses, the downstream and ImageNet spaces drift apart, which makes the model less confident and more error-prone as new classes arrive. We also observe that the KL consistency losses have an important stabilizing effect. When they are removed, the performance on Task-1 and Task-2 remains mostly unchanged as there is not much change, but the later tasks degrade. This shows that the KL terms act as a temporal regularizer, keeping the predictions from drifting too far between tasks and helping the model preserve the structure it learned earlier. Other configurations that use only a subset of these losses do provide some benefit, but they tend to emphasize one aspect of the problem at the expense of another. For example, relying only on downstream supervision biases the model toward task-specific pseudo labels, while relying only on ImageNet alignment does not fully capture the semantics of the downstream distribution. Neither approach alone is sufficient to maintain stable performance across tasks. The strongest results are consistently obtained when all components which includes alignment losses and KL consistency losses are used together proving the merit of our proposed method.

\noindent \textbf{Effect of $k$ and Auxiliary Images.} Figure~\ref{fig:acc_imgs} illustrates how accuracy changes as we vary $k$, the number of ImageNet classes retrieved per downstream class, and the number of auxiliary images sampled from each of those classes. The performance on the early tasks, Task-1 to Task-3 remains relatively stable across all settings, suggesting that the model does not require a large amount of auxiliary support when the downstream label space is still small. The impact of $k$ and image count becomes much clearer in the later tasks; Task-4 and Task-5. As more classes are introduced, the downstream data becomes more ambiguous, and the model benefits from a richer auxiliary set: increasing $k$ provides a broader semantic neighborhood, while using more images per class supplies stronger visual evidence. Together, these factors lead to consistent improvements on the more challenging later tasks, helping the model disambiguate visually similar classes in this setting.

\noindent \textbf{Qualitative Analysis using tSNE plots.} Figure ~\ref{fig:tsne} shows t-SNE visualizations of Continual CLIP (top row) and our method (bottom row) across all tasks, with ground-truth and predicted labels overlaid. Continual CLIP exhibits progressively increasing cluster overlap, with many classes collapsing into each other as tasks advance, clear evidence of representation drift and forgetting. In contrast, our method produces tighter, well-separated clusters with much closer alignment between ground truth and predictions. The structure of earlier classes is preserved even in later tasks, demonstrating stable representations and reduced confusion. These visualizations confirm that the combination of auxiliary ImageNet supervision and cross-domain alignment leads to a more robust and discriminative feature space compared to Continual CLIP.

\section{Conclusion}
\label{sec:conclusion}
We introduced an annotation free class incremental learning framework that uses world knowledge as a stable external supervisor to guide learning in the absence of labels. By aligning downstream representations with semantically related ImageNet classes and designing a cross domain training strategy that connects CLIP, DINO, and auxiliary supervision, our approach maintains coherent features across tasks while avoiding the need to store any downstream exemplars. Extensive experiments across four challenging datasets show that our method consistently outperforms CLIP based, test time adaptation based, and continual learning baselines, with especially large gains on later tasks where class confusion is highest. The qualitative analyses further demonstrate that our model retains clearer decision boundaries and suffers far less drift over time. Overall, our results highlight the value of using structured external knowledge to support continual learning when annotations are unavailable, and point towards a promising direction.
{
    \small
    \bibliographystyle{ieeenat_fullname}
    \bibliography{main}
}

% \clearpage
% \setcounter{page}{1}
\maketitlesupplementary

% Our code will be made publicly available upon acceptance for further research and reproducibility. 
This supplementary material contains additional details that we could not include in the main paper due to space constraints, including the following information.

\begin{itemize}
    \item Discussion on Methodology and Design Choices in Sec~\ref{sec:designchoice}.
    % \vspace{0.5mm}
    \item Ablation Study on Replay Samples per class is shown in  Sec~\ref{sec:replay_study}.
    % \vspace{2mm}
    \item Forgetting Measure Study in Sec~\ref{sec:forgetting}.
    % \vspace{2mm}
    \item Replacing the DINO Backbone is studied in Sec~\ref{sec:dinoalternate}.
    % \vspace{2mm}
    \item Qualitative Samples from World-Knowledge Distillation is presented in Sec~\ref{sec:semanticsimilarities}. 
    % \vspace{2mm}
    \item Algorithm of the overall proposed method is presented in Sec~\ref{sec:algorithm}.
    % \vspace{2mm}
    \item Prompts used to query LLM are presented in Sec~\ref{sec:prompts}.
\end{itemize}

\section{Discussion on Methodology and Design Choices.}
\label{sec:designchoice}
In this section we reflect on key methodological choices and address potential concerns that may arise when using ImageNet samples as auxiliary exemplars in an Annotation-free Continual Learning setting. One could wonder that whether adding such external data still qualifies as continual learning or whether it injects undesired supervision into the process. We emphasize that continual learning restricts access only to past task data, not to publicly available external sources. ImageNet samples do not belong to any task in the incremental stream and remain fixed throughout training. Their inclusion therefore does not break the sequential nature of the problem. Using them is conceptually identical to using a pretrained backbone, which every continual learning baseline already depends on. We also note that recent state of the art work such as GIFT~\cite{gift} employs synthetic ImageNet imagery generated through diffusion models and leverages auxiliary supervision even when labeled data is available. This further reinforces the motivation for incorporating auxiliary supervision in our setting, since leading continual learning methods already demonstrate the effectiveness and acceptance of such external structure. Finally, one might wonder whether the use of ImageNet makes the problem artificially easier or produces an unfair comparison. In practice, all VLM based methods already rely on massive pretrained priors, often orders of magnitude larger and richer than ImageNet. We make use of ImageNet to learn from its inherent semantic knowledge. Our use of a small, fixed set of publicly available proxy exemplars is a considerably weaker prior and provides a transparent and reproducible source of world level structure. This reflects a principled design choice: instead of relying on hidden, web scale training data or proprietary multimodal LLM supervision, we adopt a controlled and publicly accessible auxiliary source to stabilize pseudo labels without breaking the core constraints of annotation free continual learning.

\section{Ablation Study on Replay Samples per Class}
\label{sec:replay_study}
The ablation on the Oxford Pets dataset in Table~\ref{tab:replay_ablation} shows that our method is not highly sensitive to the number of auxiliary ImageNet samples used during replay. Even when varying the replay size from $k=1$ to $k=10$ samples per class, the overall performance remains stable, with average accuracy staying within a narrow band of $84$--$86\%$. This indicates that the distilled ImageNet exemplars provide strong supervisory value even in very small quantities. The model benefits consistently from the semantic structure captured in these auxiliary images, and does not rely on large replay buffers. This stands in contrast to traditional continual learning methods that require storing around $20$ real exemplars per class, which introduces storage overhead and privacy concerns. In our case, only a handful of public, non private ImageNet samples are sufficient, demonstrating both efficiency and robustness in the annotation-free setting.

\begin{table}[t]
\centering
\begin{tabular}{c c c c c c c}
\toprule
$k$ & Avg & T-1 & T-2 & T-3 & T-4 & T-5 \\
\midrule
1  & 85.67 & 92.84 & 86.11 & 85.30 & 83.76 & 80.38 \\
2  & 85.69     & 92.84    & 85.31    & 89.09    & 81.67    & 79.58 \\
5  & 84.15  & 92.84 & 85.38 & 85.30 & 80.08 & 77.19 \\
10 & 86.29  & 92.84 & 85.82 & 88.71 & 83.96 & 80.13 \\
\bottomrule
\end{tabular}
\caption{Ablation on the number of auxiliary replay samples per class ($k$) on the Oxford Pets dataset. The results show that performance remains stable across a wide range of replay sizes, indicating that our method is not highly sensitive to the exact number of ImageNet samples used.}
\label{tab:replay_ablation}
\end{table}

\section{Forgetting Measure Study}
\label{sec:forgetting}
To quantify the degradation of previously learned knowledge as new tasks arrive, we adopt the standard \emph{forgetting measure} used in continual learning. For a given task~$t$, forgetting is defined as the drop between the best accuracy ever achieved on that task and its final accuracy after completing all tasks. Formally, the forgetting for task~$t$ is computed as
\[
F(t) = \max_{k \leq T} \, a_t^{(k)} - a_t^{(T)},
\]
where $a_t^{(k)}$ denotes the accuracy on task~$t$ after training on task~$k$, and $T$ is the index of the final task. Larger values indicate more severe forgetting, while negative values imply improvement over time.

\begin{figure*}[t]
    \centering
    
    % DTD
    \begin{subfigure}{0.4\textwidth}
        \centering
        \includegraphics[width=\linewidth]{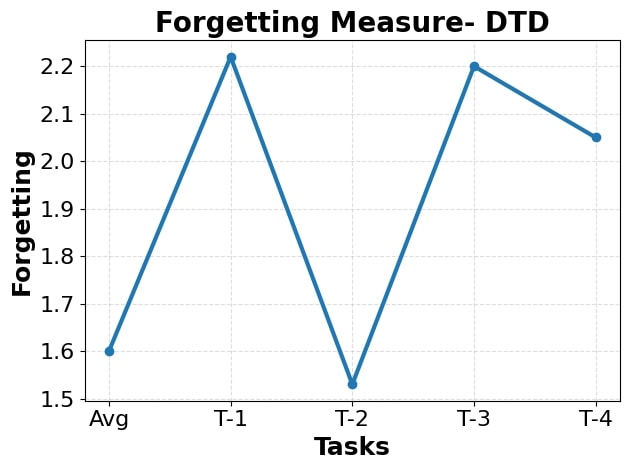}
        % \caption{DTD}
    \end{subfigure}
    \begin{subfigure}{0.4\textwidth}
        \centering
        \includegraphics[width=\linewidth]{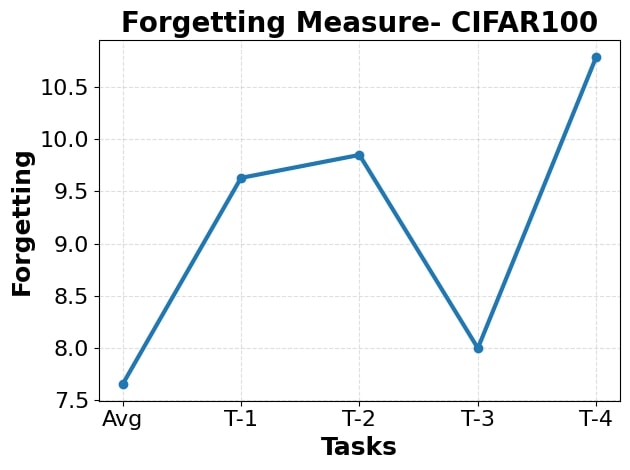}
        % \caption{CIFAR100}
    \end{subfigure}
    \begin{subfigure}{0.4\textwidth}
        \centering
        \includegraphics[width=\linewidth]{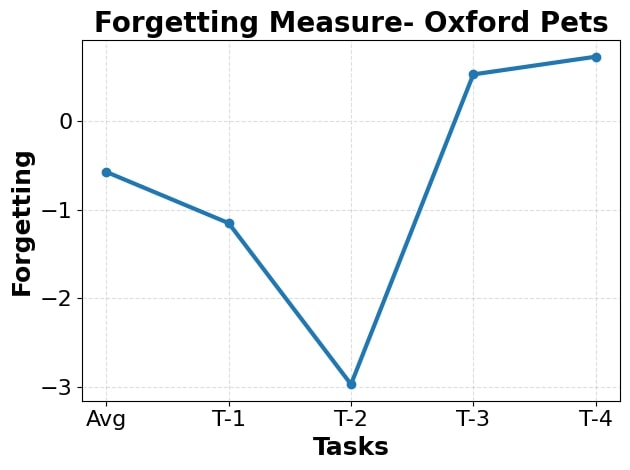}
        % \caption{Pets37}
    \end{subfigure}
    \begin{subfigure}{0.4\textwidth}
        \centering
        \includegraphics[width=\linewidth]{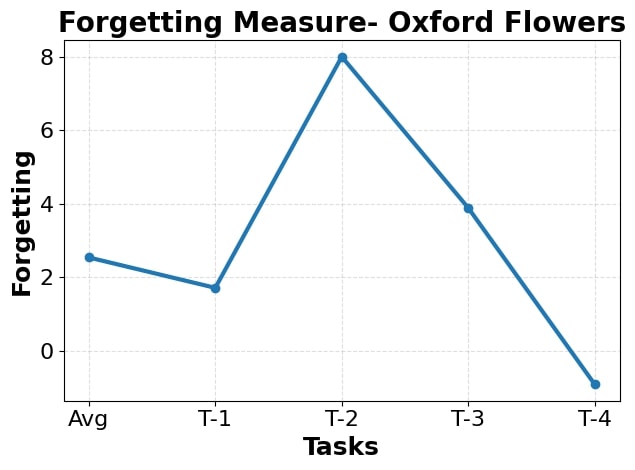}
        % \caption{Flowers}
    \end{subfigure}

    \caption{We report forgetting values for each task on DTD, CIFAR-100, Oxford Pets, and Oxford Flowers. The trends highlight dataset specific forgetting dynamics, with CIFAR-100 showing higher forgetting, Oxford Pets exhibiting negative forgetting in early tasks, and Oxford Flowers and DTD displaying moderate fluctuations.}
    \label{fig:forgetting_all}
\end{figure*}

Across the four datasets, the forgetting patterns exhibit distinct characteristics as shown Figure~\ref{fig:forgetting_all}. DTD shows mild fluctuations, with forgetting values roughly between $1.5\%$ and $2.2\%$ across tasks even with a very distinct domain gap between the distilled knowledge and the downstream task, indicating stable retention with modest performance drops. CIFAR-100 displays stronger forgetting, particularly in later tasks where values exceed~10, reflecting the dataset's higher diversity and difficulty. Despite this, our method still outperforms all baselines by a margin of $~3\%$. Oxford Pets exhibits an interesting pattern: the early tasks show negative forgetting, indicating performance gains, while the later tasks experience only mild positive forgetting, with values ranging from $-2.97\%$ to $+0.73\%$. Notably, the improvement contributed by the auxiliary data (approximately $3\%$) substantially outweighs the minimal forgetting effect (around $0.7\%$), rendering the latter practically negligible.
Oxford Flowers exhibits a mixed trajectory, where forgetting increases around task Task-2 but subsequently decreases and becomes slightly negative at Task-4, indicating partial recovery after mid sequence degradation. These trends highlight how forgetting behaves differently depending on dataset characteristics, granularity, and inter task relationships. Overall, these results show that the influence of forgetting is minimal compared to the gains achieved, confirming the robustness of our method across all datasets.

\section{Replacing the DINO Backbone}
\label{sec:dinoalternate}
The backbone ablation in Table~\ref{tab:dino_alternative} demonstrates that our method remains effective across a wide range of architectures, including lightweight models with significantly fewer parameters. Notably, when replacing the DINO 16-bit backbone (86M params) with the much smaller ViT-Tiny model (only 5.7M params), the performance remains remarkably stable. Despite being more than an order of magnitude smaller, ViT-Tiny achieves an overall average of $69.97\%$, outperforming the best prior SOTA by a margin of $+4.60\%$. Furthermore, the dataset-wise improvements remain consistent, showing strong gains on Oxford Flowers and Oxford Pets, and competitive results on DTD and CIFAR-100. This indicates that the benefits of our auxiliary ImageNet-based distilled supervision are not tied to a specific architecture capacity; instead, the method generalizes reliably even with compact models. The strong performance of such a lightweight backbone highlights the efficiency and robustness of our approach, suggesting that the gains stem primarily from the quality of auxiliary supervision rather than the scale of the model. We also show a performance gain of $+7.47\%$ on DINO 8-bit backbone (21M params) which is slightly bigger than ViT-Tiny, making the method more robust across different backbones. 

\begin{table*}[t]
\centering
\setlength{\tabcolsep}{4pt}
\begin{tabular}{llccccccc}
\toprule
Backbone & Dataset & Overall Avg & Avg & T-1 & T-2 & T-3 & T-4 & T-5 \\
\midrule
\multirow{4}{*}
% {Best-performing SOTA (RAPF-ECCV'24)}
    {
  \parbox{2.8cm}{\centering
   Best-performing SOTA\\
    \small (RAPF (ECCV'24))
  }
}
    & DTD            & \multirow{4}{*}     
    {65.37} & 53.66 & 79.17 & 57.64 & 47.32 & 43.35 & 40.84 \\
    & CIFAR100       & &
    58.04 & 82.89 & 63.79 & 54.06 & 46.58 & 42.92 \\
    & Oxford Pets & &
    \underline{80.74} & 92.69 & 79.09 & 78.42 & 75.81 & 77.71 \\
    & Oxford Flowers & &
    \underline{69.03} & 91.97 & 73.68 & 61.69 & 60.32 & 57.53 \\
\midrule
\multirow{4}{*}{
  \parbox{2.8cm}{\centering
    DINO 16-bit (86M)\\
    \small (CrossWorld-CL from Table~[2] of main paper)
  }
}
    & DTD            & \multirow{4}{*}{71.912 \textbf{\textcolor{blue}{(+6.54)}}} & 62.658 & 84.17 & 74.31 & 57.85 & 51.10 & 45.86 \\
    & CIFAR100       &                        & 67.440 & 93.11 & 79.85 & 66.60 & 51.45 & 46.19 \\
    & Oxford Pets    &                        & 86.340 & 93.12 & 85.82 & 88.71 & 83.96 & 80.13 \\
    & Oxford Flowers &                        & 71.210 & 94.02 & 80.54 & 63.95 & 60.86 & 56.68 \\

\midrule
\multirow{4}{*}{ViT-Tiny (5.7M)} 
    & DTD            & \multirow{4}{*}{69.972 \textcolor{blue}{\textbf{(+4.60)}}} & 62.656 & 80.28 & 72.64 & 58.81 & 53.80 & 47.75 \\
    & CIFAR100       &                        & 61.922 & 89.53 & 75.46 & 62.70 & 48.47 & 33.45 \\
    & Oxford Pets    &                        & 83.550 & 87.54 & 82.56 & 87.17 & 81.05 & 79.45 \\
    & Oxford Flowers &                        & 71.760 & 94.59 & 81.83 & 64.65 & 60.38 & 57.37 \\

\midrule
\multirow{4}{*}{DINO 8-bit (21M)} 
    & DTD            & \multirow{4}{*}{72.8485\textcolor{blue}{\textbf{(+7.47)}}} & 65.106 & 84.44 & 76.67 & 61.02 & 53.87 & 49.58 \\
    & CIFAR100       &                         & 69.872 & 92.53 & 80.13 & 68.85 & 58.85 & 49.00 \\
    & Oxford Pets    &                         & 83.530 & 89.54 & 84.80 & 88.60 & 78.06 & 76.67 \\
    & Oxford Flowers &                         & 72.886 & 94.87 & 83.26 & 67.52 & 61.29 & 57.49 \\
\bottomrule
\end{tabular}
\caption{Replacing DINO with lighter backbones such as ViT-Tiny or 8-bit DINO still provides strong and consistent performance, indicating backbone independence. The table reports task-wise results (Avg and Task-1 to Task-5) and the Overall Avg for each backbone. The gains in \textcolor{blue}{blue} are reported against the best performing SOTA method.}
\label{tab:dino_alternative}
\end{table*}

\section{Qualitative Samples from World-Knowledge Distillation}
\label{sec:semanticsimilarities}
\begin{figure*}[t]
    \centering
    
    % Row 1
    \begin{subfigure}{0.47\textwidth}
        \centering
        \includegraphics[width=\linewidth]{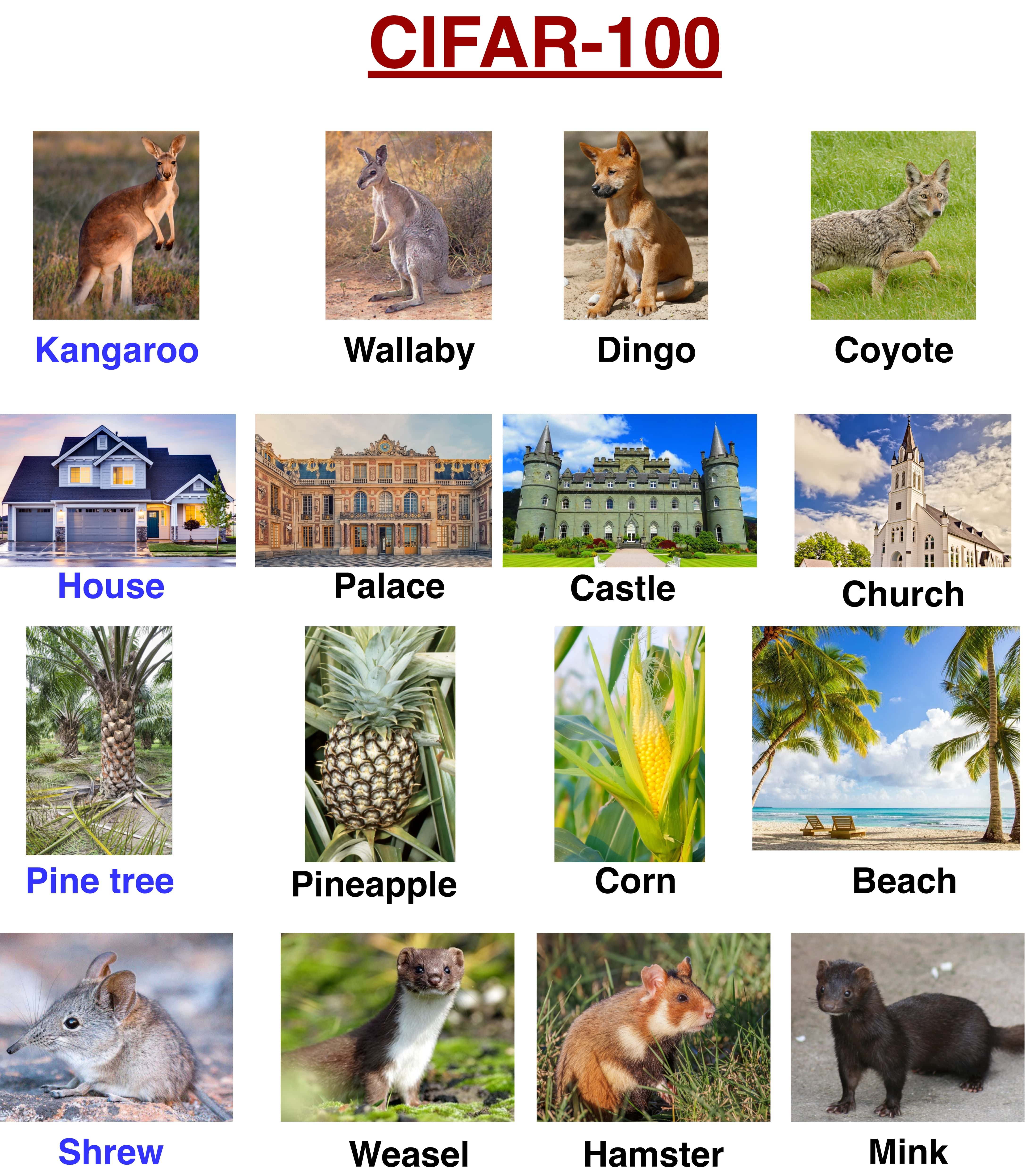}
        % \caption{CIFAR-100}
    \end{subfigure}
    \hfill
    \begin{subfigure}{0.47\textwidth}
        \centering
        \includegraphics[width=\linewidth]{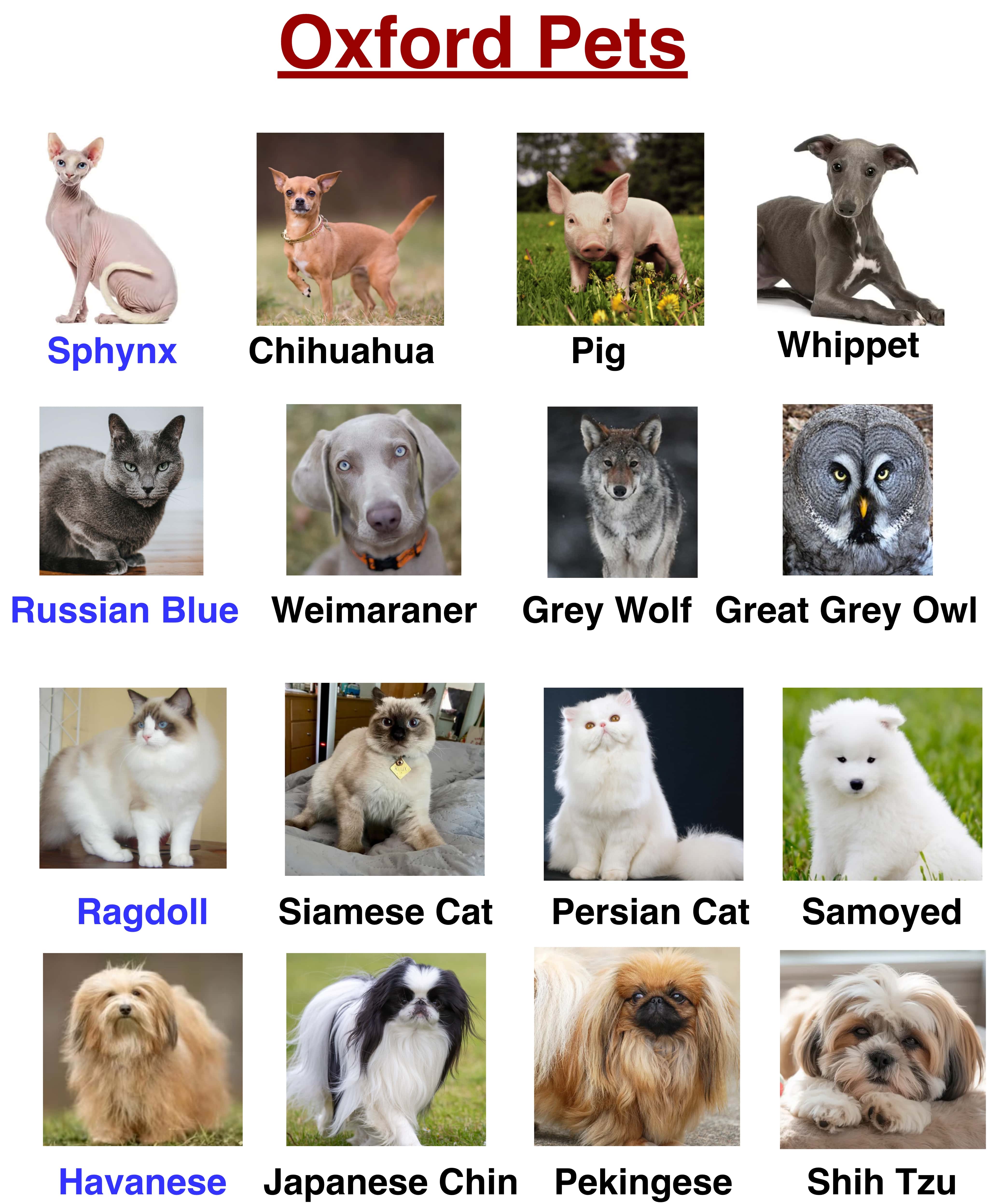}
        % \caption{Oxford Pets}
    \end{subfigure}
    % Row 2
    \begin{subfigure}{0.47\textwidth}
        \centering
        \includegraphics[width=\linewidth]{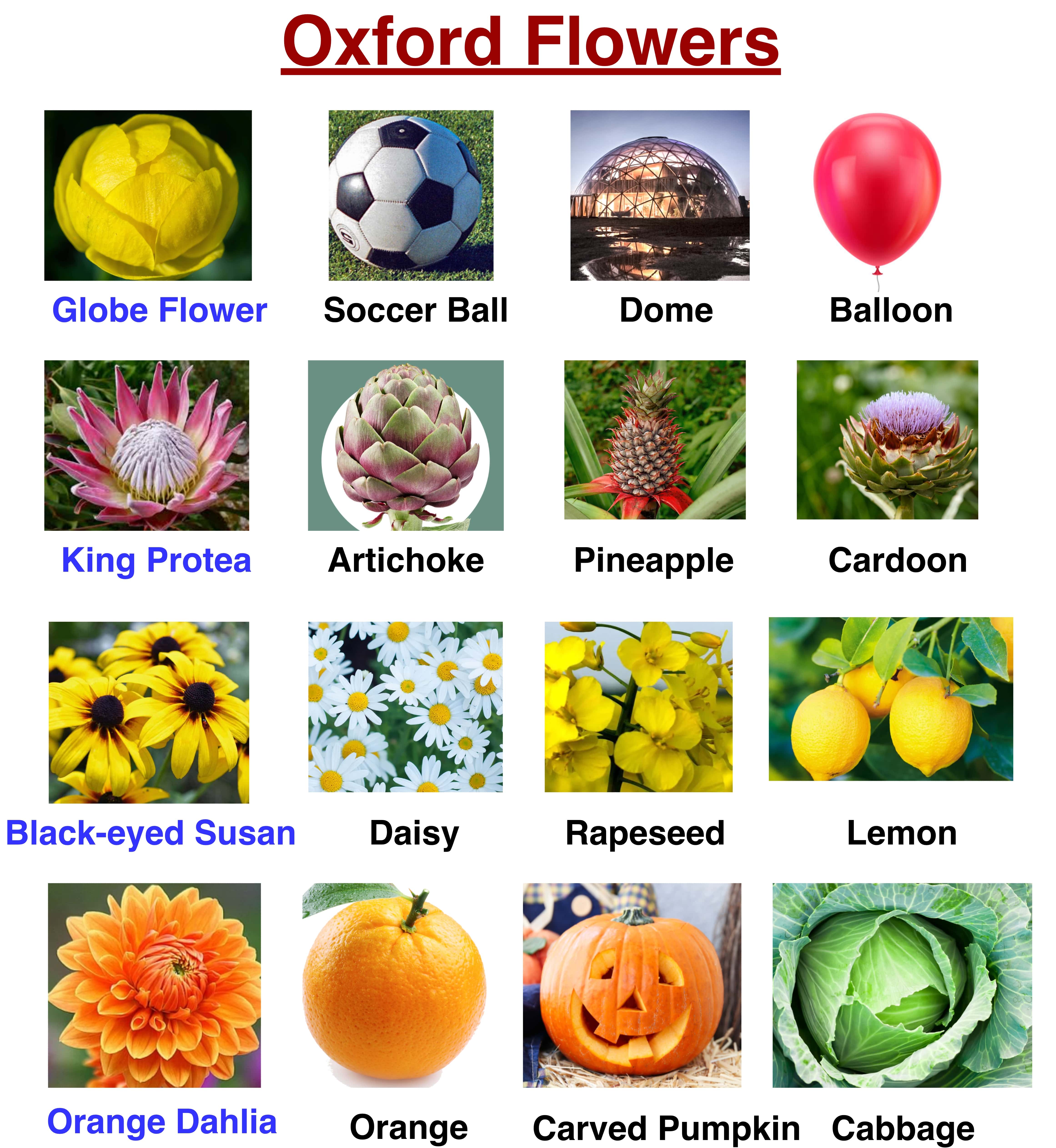}
        % \caption{Oxford Flowers}
    \end{subfigure}
    \hfill
    \begin{subfigure}{0.47\textwidth}
        \centering
        \includegraphics[width=\linewidth]{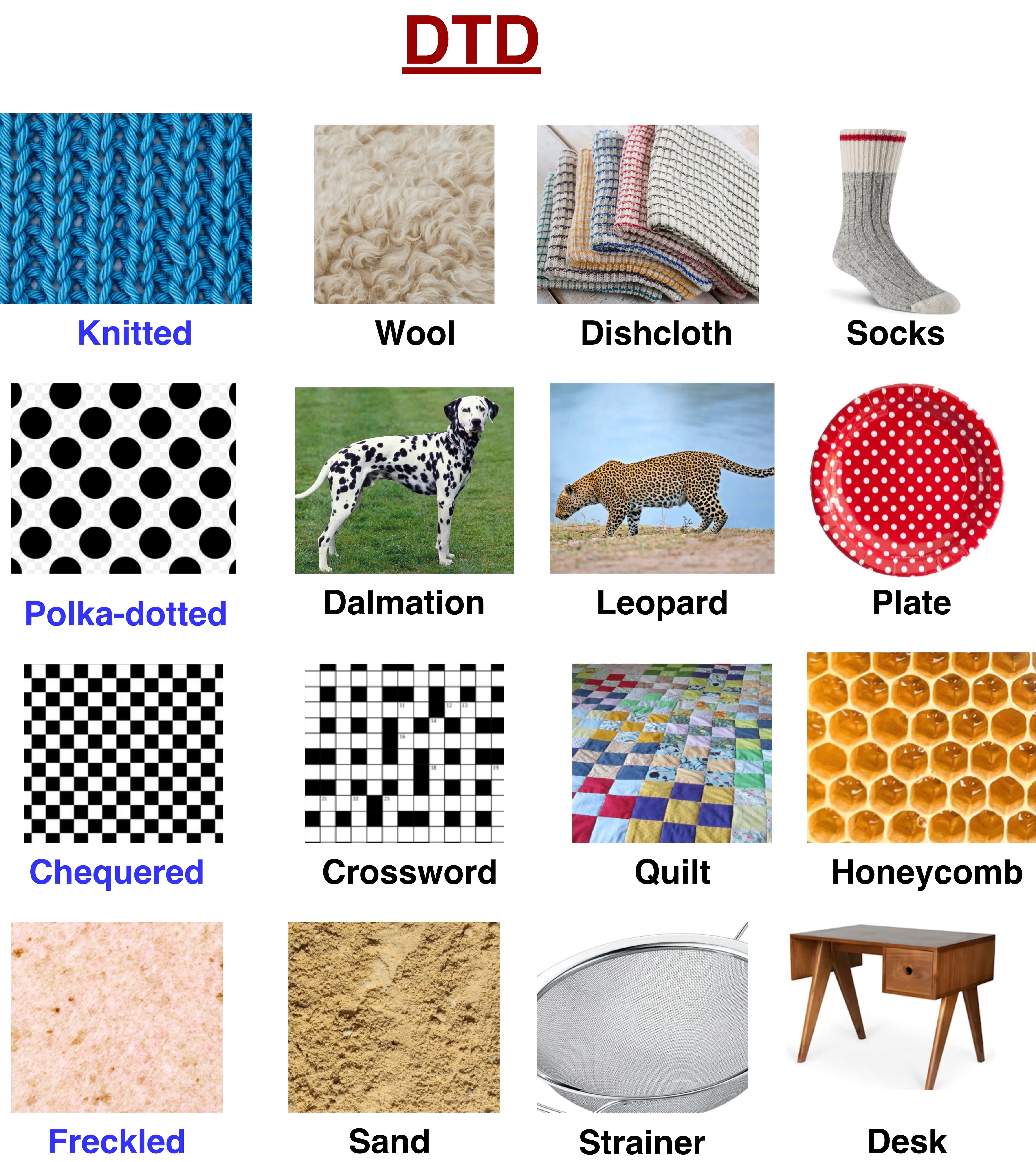}
        % \caption{DTD}
    \end{subfigure}
    \caption{Qualitative examples of distilled world knowledge. For each downstream class (in \textcolor{blue}{blue}), we show semantically related ImageNet images retrieved through our distillation step. These examples highlight how auxiliary world knowledge provides meaningful visual cues that align with downstream classes, enabling privacy-safe and effective guidance for annotation-free continual learning.}
    \label{fig:world_knowledge}
    \vspace{10mm}
\end{figure*}

The examples in Figure~\ref{fig:world_knowledge} illustrate how world knowledge from ImageNet can be distilled to support downstream classes in an Annotation-Free Continual Learning paradigm. For each downstream class (shown in \textcolor{blue}{blue}), we retrieve a small set of semantically related ImageNet classes, displayed adjacent to it. These retrieved concepts consistently reveal meaningful visual or semantic correspondences: for instance, CIFAR--100's \textcolor{blue}{Kangaroo} retrieves ImageNet classes containing similar animal poses and textures, while Oxford Flowers' \textcolor{blue}{Globe Flower} retrieves spherical or patterned objects such as \emph{Soccer Ball} and \emph{Balloon}, capturing strong shape-based similarity. Likewise, in DTD, texture classes such as \textcolor{blue}{Knitted} or \textcolor{blue}{Polka-dotted} retrieve ImageNet samples exhibiting similar structural repetition, color distribution, and geometric regularity. These relationships highlight that ImageNet contains a rich semantic prior that can be repurposed, even without explicit labels as auxiliary supervision for a wide range of downstream categories.

The retrieved samples thus act as proxy exemplars that encode the high-level concepts associated with each downstream class, providing the model with stable semantic anchors across tasks. Because they capture discriminative properties such as texture, shape, object geometry, or fine-grained appearance cues, they help reinforce and stabilize representations during continual learning, particularly when the downstream data arrive unlabeled. Importantly, these ImageNet exemplars remove the need to store any real user or task-specific images, addressing long-standing concerns around privacy and data retention in continual learning. By leveraging publicly available world-knowledge images that are semantically mapped to downstream classes, our method delivers a principled and privacy-preserving replay mechanism while still achieving substantial performance improvements across all tasks.

% \FloatBarrier

\section{Algorithm}
\label{sec:algorithm}

\textbf{Algorithm-1} provides an overview of our Annotation-Free Continual Learning pipeline, where each stage contributes a complementary form of structure to stabilize learning across tasks. At a high level, the algorithm first retrieves relevant world knowledge from ImageNet, expands the downstream task knowledge semantically using an LLM, aligns visual and textual spaces using dual supervision by mapping DINO to CLIP space, further enforces cross-domain consistency through prompt-guided alignment, and finally selects a compact replay set to guide future tasks. Each stage has a clear intuition: Stage-1 identifies ImageNet classes that are semantically closest to the downstream labels, giving the model a ``world-knowledge prior'' that anchors its representations \textbf{(Algorithm-2)}. Stage-2 enriches downstream text embeddings by using an LLM to provide richer, more descriptive semantics \textbf{(Algorithm-3)}, helping the model reason at a concept level rather than from short label tokens. Stage-3 then aligns both downstream and auxiliary images from DINO visual space to CLIP text spaces \textbf{(Algorithm-4)}, ensuring that visual and linguistic semantics reinforce each other by training two adapters. The trained DINO adapters act as a pseudo-labeler to train the prompted CLIP model in next stage. The reason DINO is chosen because of it's ability to capture object-centric features that later help the model in cross-domain alignment. Stage-4 enforces cross-domain alignment between CLIP and DINO, allowing both encoders to agree on a shared semantic space and preventing drift as new tasks arrive \textbf{(Algorithm-5)}. This way the prompted CLIP model learns a generic representation using DINO and the auxiliary data with the help of all the alignment losses.  Finally, Stage-5 distills the auxiliary pool into a small set of high-confidence exemplars using prompted CLIP model scores \textbf{(Algorithm-6)}, forming a replay memory that is privacy-safe, compact, and semantically meaningful. Together, these stages create a stable learning framework where external world knowledge reduces noise in pseudo labels, alignment losses tie representations together across domains, and replay ensures
long-term retention across tasks. We also clarify that the auxiliary ImageNet data is introduced only from Task-2 onward. Since Task-1 contains very few classes, we avoid adding auxiliary samples at this stage to prevent the model from being confused early in training. We also clarify that we don't use any task-id during inference.

\section{Prompts used to query LLM}
\label{sec:prompts}
In Stage~2 we query an large language model (GPT 3.5) with class names to obtain richer semantic descriptions of each downstream category. For every class label, we instantiate a small set of dataset specific prompt templates and send only the text label (no images) to the LLM. The returned descriptions include attributes, shapes, materials, typical contexts, and synonyms. We then concatenate these descriptions with
the original label and encode the resulting text using the CLIP text encoder to form the task aware embeddings $t_D$. The same descriptions are also used to refine the mapping between downstream labels and candidate ImageNet classes. In this way, Stage~2 injects high level semantic knowledge into the text space while remaining annotation free: the LLM is never asked to label images, but only to
explain what each class usually looks like in a realistic photograph that matches the dataset domain.

\medskip
\noindent\textbf{DTD (textures) prompts.}
\begin{itemize}
    \item \texttt{Describe a close up texture of \{category\}.}
    \item \texttt{How does the \{category\} texture usually look in a photograph?}
    \item \texttt{What visual pattern and material properties define the \{category\} texture?}
    \item \texttt{How can you recognize \{category\} from a small image patch?}
    \item \texttt{Describe a photo that clearly shows the \{category\} texture.}
\end{itemize}

\noindent\textbf{CIFAR-100 prompts.}
\begin{itemize}
    \item \texttt{Describe a natural photo of \{category\}.}
    \item \texttt{How does \{category\} usually appear in a real world image?}
    \item \texttt{What visual details help you identify \{category\} in a picture?}
    \item \texttt{Describe a typical scene that contains \{category\}.}
    \item \texttt{How can you recognize \{category\} when looking at a single image?}
\end{itemize}

\noindent\textbf{Oxford Pets prompts.}
\begin{itemize}
    \item \texttt{Describe a portrait photo of a \{category\} pet.}
    \item \texttt{How does a \{category\} cat or dog usually look in a photo?}
    \item \texttt{What facial features and fur patterns are typical for a \{category\}?}
    \item \texttt{How can you identify a \{category\} in a pet photograph?}
    \item \texttt{Describe a clear photo that shows a \{category\} from the front.}
\end{itemize}

\noindent\textbf{Oxford Flowers prompts.}
\begin{itemize}
    \item \texttt{Describe a close up photo of a \{category\} flower.}
    \item \texttt{What are the typical color, shape, and structure of a \{category\} blossom?}
    \item \texttt{How does a garden photo of \{category\} flowers usually look?}
    \item \texttt{How can you recognize a \{category\} in a photograph of plants?}
    \item \texttt{Describe a detailed image that focuses on a single \{category\} bloom.}
\end{itemize}

\begin{algorithm}[t]
\caption{CrossWorld-CL for Annotation Free Continual Learning Paradigm}
\label{alg:overall_afcl}
\begin{algorithmic}[1]
\Require Stream of tasks $\{D^{t}\}_{t=1}^{T}$ where $D^{t} = \{X_D^{t}, Y_D^{t}\}$ 
\Require ImageNet auxiliary pool $\mathcal{X}_{I}$; CLIP encoders $f_{I}$ (image), $f_{t}$ (text); DINO backbone $f_{D}$
\Require LLM; replay budget $k$ (top $k$ samples per class)
\Ensure Trained adapters $\phi_{D}, \phi_{I}$, llm\_prompts, and replay memory $\mathcal{A}_{R}$

\State Initialize adapters $\phi_{D}, \phi_{I}$ and prompts randomly
\State Initialize global replay memory $\mathcal{A}_{R} \gets \emptyset$

\For{$t = 1$ to $T$}
    \Statex \textbf{Stage 1: Task Aware World Knowledge Distillation}
    \State $\mathcal{A}^{t}$
     $\gets$ World Knowledge Distillation ({$Y_D^{t}, \mathcal{X}_I, \mathcal{Y}_I, f_{t}$})
    
    \Statex \textbf{Stage 2: Semantic Expansion through LLM}
    \State $t_{D}, t_{I} \gets$ 
    Semantic Expansion ({$\mathcal{Y}_D^t, LLM, \mathcal{A}^t, f_{t}$, llm\_prompts})
    
    \Statex \textbf{Stage 3: Dual Supervised Visual Semantic Alignment}
    \State $\phi_{D}, \phi_{I} \gets$
    Dual Supervised Alignment ({$\mathcal{X}_D^{t}, A^{t}, t_{D}, f_{I}, f_{D}, \phi_{D}, \phi_{I}$})
    
    \Statex \textbf{Stage 4: Prompt Guided Cross Domain Alignment}
    \State $f_{I_p}, t_D, t_I \gets$
    Cross Domain Alignment ({$\mathcal{X}_D^{t}, \mathcal{A}^{t}, t_{D}, t_{I}, f_{I_p}, f_D, \phi_{D}, \phi_{I}$})
    
    \Statex \textbf{Stage 5: Replay Strategy}
    \State $\mathcal{A}_{R} \gets$ Build Replay ({$\mathcal{A}^{t}, f_{I_p}, t_D$})
    \State $A^{R} \gets A^{R} \cup A^{t}_{R}$
\EndFor

\State \Return $t_{D}, f_{I_p}$ 

\Statex
\Statex \textbf{Inference on test data}
\For{each test image $x \in \mathcal{T}_{test}^t$}
    \State $v \gets \mathrm{norm}\!\big(f_{I_p}(x)\big)$
    \State \ForAll{$c \in \mathcal{C}$} \Comment{$\mathcal{C}$: all downstream classes seen so far}
              \State \hspace{4mm} $s[c] \gets \cos\big(v, t_{D}[c]\big)$
          \EndFor
    \State $\hat{y}(x) \gets \arg\max_{c \in \mathcal{C}} s[c]$
\EndFor
\State \Return predictions $\hat{y}(x)$ for all $x \in \mathcal{T}_{test}^t$

\end{algorithmic}
\end{algorithm}

\begin{algorithm}[t]
\caption{Stage 1: World Knowledge Distillation}
\label{alg:stage1_world}
\begin{algorithmic}[1]
\Function{World Knowledge Distillation} 
 {$Y_D^{t}, \mathcal{X}_I, \mathcal{Y}_I, f_{t}$}
    \Require Downstream class names $Y_D^{t}$; ImageNet images $\mathcal{X}_I$ and labels $\mathcal{Y}_I$; text encoder $f_t$
    \Ensure Task specific auxiliary pool $\mathcal{A}^{t}$

    \State $\mathcal{A}^{t} \gets \emptyset$
    \Statex \Comment{Encode downstream and ImageNet label texts}
    \For{each $c \in Y_D^{t}$}
        \State $t_D^{\text{raw}}[c] \gets \mathrm{norm}(f_t(c))$
    \EndFor
    \For{each $u \in \mathcal{Y}_I$}
        \State $t_I^{\text{all}}[u] \gets \mathrm{norm}(f_t(u))$
    \EndFor

    \Statex \Comment{Find ImageNet labels closest to each downstream label}
    \For{each $c \in Y_D^{t}$}
        \State $\mathcal{S}_c \gets \mathrm{Topk}_{u \in \mathcal{Y}_I}
            \big(\cos(t_D^{\text{raw}}[c], t_I^{\text{all}}[u])\big)$
        \For{each $u \in \mathcal{S}_c$}
            \State $\mathcal{A}^{t} \gets \mathcal{A}^{t} \cup 
                   \{x \in \mathcal{X}_I \mid \text{label}(x) = u\}$
        \EndFor
    \EndFor

    \State \Return $\mathcal{A}^{t}$
\EndFunction
\end{algorithmic}
\end{algorithm}

\begin{algorithm}[t]
\caption{Stage 2: Semantic Expansion through LLM}
\label{alg:stage2_semantic}
\begin{algorithmic}[1]
\Function{Semantic Expansion}{$\mathcal{Y}_D^{t}, \text{LLM}, \mathcal{A}^t, f_{t}, llm\_prompts$}
    \Require $\mathcal{Y}_D^{t}$; LLM; $\mathcal{A}^t$; text encoder $f_t$; llm\_prompts
    \Ensure Downstream and ImageNet text embeddings $t_D, t_I$

    \State $t_D \gets \emptyset$, \quad $t_I \gets \emptyset$

    \For{each class $c \in \mathcal{Y}_D^{t}$}
        \State $\text{prompt}_c \gets$ build prompt from $c$ and llm\_prompts
        \State $\{\text{desc}_{c}^{(1)}, \ldots, \text{desc}_{c}^{(m)}\} \gets$ query LLM with $\text{prompt}_c$

        \Statex \hspace{2mm} \Comment{Compute multiple text embeddings and average them}
        \State Initialize list $\mathcal{E}_c \gets \emptyset$
        \For{$j = 1$ to $m$}
            \State $\mathcal{E}_c \gets \mathcal{E}_c \cup 
                \left\{\mathrm{norm}\!\left(f_t(\text{desc}_{c}^{(j)})\right)\right\}$
        \EndFor
        \State $t_D[c] \gets \frac{1}{m} \sum_{e \in \mathcal{E}_c} e$

        \Statex \hspace{2mm} \Comment{Collect ImageNet text embeddings for selected classes in $\mathcal{A}^t$}
        \For{each $u \in \mathcal{Y}_I$}
            \If{$u$ in $\mathcal{A}^t$}
                \State $t_I[u] \gets \mathrm{norm}\!\left(f_t(u)\right)$
            \EndIf
        \EndFor
    \EndFor

\State \Return $t_D, t_I$
\EndFunction
\end{algorithmic}
\end{algorithm}

\begin{algorithm}[t]
\caption{Stage 3: Dual Supervised Visual Semantic Alignment}
\label{alg:stage3_dual_alignment}
\begin{algorithmic}[1]
\Function{Dual Supervised Alignment} {$\mathcal{X}_D^{t}, A^{t}, t_{D}, f_{I}, f_{D}, \phi_{D}, \phi_{I}$}
\Require Current task images $X_D^{t}$; auxiliary dataset $\mathcal{A}^{t}$
\Require Downstream and ImageNet Text embeddings $t_{D}$, $t_{I}^{t}$
\Require Encoders $f_{D}$ (DINO), $f_{I}$ (CLIP image); adapters $\phi_{D}, \phi_{I}$
\Ensure Updated adapters $\phi_{D}, \phi_{I}$

\vspace{1mm}
\Statex \textbf{Step 1: Obtain pseudo labels for downstream and replay images}
\For{each $x \in X_{\text{D}}^t$}
    \State $z \gets \mathrm{norm}\!\big(\phi_{D}(f_{D}(x))\big)$
    \State $\hat{y}_{D_i}^t \gets \displaystyle
           \arg\max_{c} \cos\big(z, t_{D}[c]\big)$
\EndFor

\vspace{1mm}
\Statex \textbf{Step 2: Train adapters with dual supervision}
\For{each training step}
    \State Sample mini batches $B_{\text{D}} \subset X_{\text{D}}^t$ and $B_{\text{I}} \subset \mathcal{A}^{t}$
    
    \Statex \Comment{Aligning Downstream images with downstream text}
    % \For{each $x \in B_{\text{down}}$}
    \State $z_{D} \gets \mathrm{norm}\!\big(\phi_{D}(f_{D}(B_D))\big)$
    \State $L_{map_1} = \mathcal{L}_{\text{CE}}\big(z_{D}, \hat{y}_{D_i}^t)$

    \Statex \Comment{Aligning Auxiliary ImageNet images with ImageNet text}
    \State $z_{I} \gets \mathrm{norm}\!\big(\phi_{I}(f_{D}(B_I))\big)$
    \State $L_{map_2} = \mathcal{L}_{\text{CE}}\big(z_{I}, y_{I_i}^t)$
    
    \State $L_{map} \gets L_{map_1} + L_{map_2}$
    
    \State Update $\phi_{D}, \phi_{I}$ using gradient of $L_{map}$
\EndFor

\State \Return $\phi_{D}, \phi_{I}$
\EndFunction
\end{algorithmic}
\end{algorithm}

\begin{algorithm}[t]
\caption{Stage 4: Prompt Guided Cross Domain Alignment}
\label{alg:stage4_cda}
\begin{algorithmic}[1]
\Function{Cross Domain Alignment}{$\mathcal{X}_D^{t}, \mathcal{A}^{t}, t_D, t_I, f_{I_p}, f_D, \phi_D, \phi_I$}
    \Require Downstream images $\mathcal{X}_D^{t}$; auxiliary pool $\mathcal{A}^{t}$
    \Require Text embeddings $t_D, t_I$; prompted CLIP $f_{I_p}$; DINO $f_D$; adapters $\phi_D, \phi_I$
    \Ensure Updated $f_{I_p}$, $t_D, t_I$
    
    \For{each training step}
        \State Sample mini batch $B_D, B_I$
        \State $L_{\text{align}} \gets 0$

        % \For{each $x \in B$}
        \State $C_{DD} \gets t_D(f_{I_p}(\mathrm{aug}(B_D)))$
        \State $C_{ID} \gets t_D(f_{I_p}(\mathrm{aug}(B_I)))$
        \State $C_{DI} \gets t_I(f_{I_p}(\mathrm{aug}(B_D)))$
        \State $C_{II} \gets t_I(f_{I_p}(\mathrm{aug}(B_I)))$

        \State $D_{DD} \gets \phi_D(f_{D}(\mathrm{aug}(B_D)))$
        \State $D_{ID} \gets \phi_D(f_{D}(\mathrm{aug}(B_I)))$
        \State $D_{DI} \gets \phi_I(f_{D}(\mathrm{aug}(B_D)))$

        \Statex \Comment{Compute cross-domain alignment losses}
        \[
        \mathcal{L}_{DD} = 
        \mathcal{L}_{\mathrm{CE}}\!\left(
            C_{DD}, D_{DD}
        \right)
        \]
        \[
        \mathcal{L}_{II} = 
        \mathcal{L}_{\mathrm{CE}}\!\left(
            C_{II}, y_{I_i}^t
        \right)
        \]
        \[
        \mathcal{L}_{ID} = 
        \mathcal{L}_{\mathrm{CE}}\!\left(
            C_{ID}, D_{ID}
        \right)
        \]
        \[
        \mathcal{L}_{DI} = 
        \mathcal{L}_{\mathrm{CE}}\!\left(
            C_{DI}, D_{DI}
        \right)
        \]

        \Statex \Comment{Combine them}
        \[
        L_{\text{align}} \mathrel{+}= 
        \mathcal{L}_{DD}
        + \lambda_1 \mathcal{L}_{II}
        + \lambda_2 \mathcal{L}_{ID}
        + \lambda_3 \mathcal{L}_{DI}
        \]

        \If{$t > 1$}
            \State $\mathcal{L}_{\mathrm{KD}} \gets
            \mathrm{KL}\!\left(
                \mathrm{Softmax}\!\left(\frac{t_D^{t-1}}{\tau}\right)
                \,\Big\|\,
                \mathrm{Softmax}\!\left(\frac{t_D^{t}}{\tau}\right)
            \right)
            +
            \mathrm{KL}\!\left(
                \mathrm{Softmax}\!\left(\frac{t_I^{t-1}}{\tau}\right)
                \,\Big\|\,
                \mathrm{Softmax}\!\left(\frac{t_I^{t}}{\tau}\right)
            \right)$
            \State $\mathcal{L}_{\text{align}} \mathrel{+}= 
            \lambda_{\mathrm{KD}}\, \mathcal{L}_{\mathrm{KD}}$
        \EndIf

        \State Update $f_{I_p}, t_D, t_I$  using $\nabla L_{\text{align}}$
    \EndFor

    \State \Return $f_{I_p}, t_D, t_I$
\EndFunction
\end{algorithmic}
\end{algorithm}

\begin{algorithm}[t]
\caption{Stage 5: Replay Strategy}
\label{alg:stage5_replay}
\begin{algorithmic}[1]
\Function{Build Replay}{$\mathcal{A}^{t}, f_{I_p}, t_D$}
    \Require Auxiliary pool $\mathcal{A}^{t}$; prompted CLIP $f_{I_p}$; downstream text embeddings $t_D$
    \Ensure Class balanced replay set $\mathcal{A}_{R}^{t}$

    \State Initialize map $\mathcal{G}[c] \gets$ empty list for all classes $c$
    \For{each image $a \in \mathcal{A}^{t}$}
        \State $v_p(a) \gets \mathrm{norm}(f_{I_p}(a))$
        \State $c^* \gets \arg\max_{c} \cos\big(v_p(a), t_D[c]\big)$
        \State $s(a) \gets \max_{c} \cos\big(v_p(a), t_D[c]\big)$
        \State append $(a, s(a))$ to $\mathcal{G}[c^*]$
    \EndFor

    \State $\mathcal{A}_{R}^{t} \gets \emptyset$
    \For{each class $c$}
        \State Sort $\mathcal{G}[c]$ in descending order of $s(a)$
        \State Add top $k$ images in $\mathcal{G}[c]$ to $\mathcal{A}_{R}^{t}$
    \EndFor

    \State \Return $\mathcal{A}_{R}^{t}$
\EndFunction
\end{algorithmic}
\end{algorithm}

\end{document}